%% file: main.tex
\documentclass[times,twocolumn,final,authoryear]{elsarticle}

\usepackage{ycviu}
\usepackage{framed,multirow}

\usepackage{amssymb}
\usepackage{latexsym}
\usepackage{balance}
\usepackage{url}
\usepackage{xcolor}
\definecolor{newcolor}{rgb}{.8,.349,.1}

\journal{Computer Vision and Image Understanding}

\newcommand{\etal} {\textit{et.al.}}
\usepackage{subfigure}
\usepackage{subfig}

\begin{document}

\thispagestyle{empty}
                                                             
\begin{frontmatter}

\title{Scalable learning for bridging the species gap in image-based plant phenotyping}

\author[1]{Daniel \snm{Ward}\corref{cor1}} 
\author[1]{Peyman \snm{Moghadam}\corref{cor2}}

\cortext[cor2]{Corresponding author: 
  Tel.: +61-7-3327-4601;}
\ead{Peyman.Moghadam@data61.csiro.au}

\address[1]{Robotics and Autonomous Systems Group\newline The Commonwealth Scientific and Industrial Research Organisation (CSIRO), Data61, Brisbane, Australia}

\received{1 May 2013}
\finalform{10 May 2013}
\accepted{13 May 2013}
\availableonline{15 May 2013}
\communicated{S. Sarkar}

\begin{abstract}
\input{chapters/0_abstract.tex}

\end{abstract}


\end{frontmatter}

\input{chapters/1_introduction.tex}

\input{chapters/2_relatedWork.tex}
\input{chapters/3_method.tex}

\input{chapters/4-1_Exp.tex}
\input{chapters/4_results.tex}
\input{chapters/5_discussion.tex}

\balance

\bibliographystyle{model2-names}
\bibliography{main}

\clearpage

\setcounter{section}{0}
\setcounter{figure}{0}
\setcounter{page}{1}
\section*{Supplementary Material}

\input{chapters/8_appendix.tex}
\input{chapters/7_visuals.tex}

\end{document}

%% file: chapters/0_abstract.tex
The traditional paradigm of applying deep learning -collect, annotate and train on data- is not applicable to image-based plant phenotyping. Data collection involves the growth of many physical samples, imaging them at multiple growth stages and finally manually annotating each image. This process is error-prone, expensive, time consuming and often requires specialised equipment. Almost 400,000 different plant species exist across the world. Each varying greatly in appearance, geometry and structure, a \textit{species gap} exists between the domain of each plant species. The performance of a model is not generalisable and may not transfer to images of an unseen plant species. With the costs of data collection and number of plant species, it is not tractable to apply deep learning to the automation of plant phenotyping measurements. Hence, training using synthetic data is effective as the cost of data collection and annotation is free.
We investigate the use of synthetic data for image-based plant phenotyping. Our conclusions and released data are applicable to the measurement of phenotypic traits including plant area, leaf count, leaf area and shape. In this paper, we validate our proposed approach on leaf instance segmentation for the measurement of leaf area.
We study multiple synthetic data training regimes using Mask-RCNN when few or no annotated real data is available. We also present \textit{UPGen}: a Universal Plant Generator for bridging the \textit{species gap}.
\textit{UPGen} leverages domain randomisation to produce widely distributed data samples and models stochastic biological variation. A model trained on our synthetic dataset traverses the \textit{domain} and \textit{species gaps}. In validating \textit{UPGen}, the relationship between different data parameters and their effects on leaf segmentation performance is investigated.
Imitating a plant phenotyping facility processing a new plant species, our methods outperform standard practices, such as transfer learning from publicly available plant data, by 26.6\% and 51.46\% on two unseen plant species respectively. 
We benchmark \textit{UPGen} by using it to compete in the CVPPP Leaf Segmentation Challenge. Generalising across multiple plant species, our method achieved state-of-the-art performance scoring a mean of 88\% across A1-4 test datasets. Our synthetic dataset and pretrained model are available at \url{https://csiro-robotics.github.io/UPGen_Webpage/}.

%% file: chapters/1_introduction.tex
\section{Introduction}\label{sec_introduction}

 \begin{figure*}
	\centering
	\includegraphics[width=0.75\textwidth]{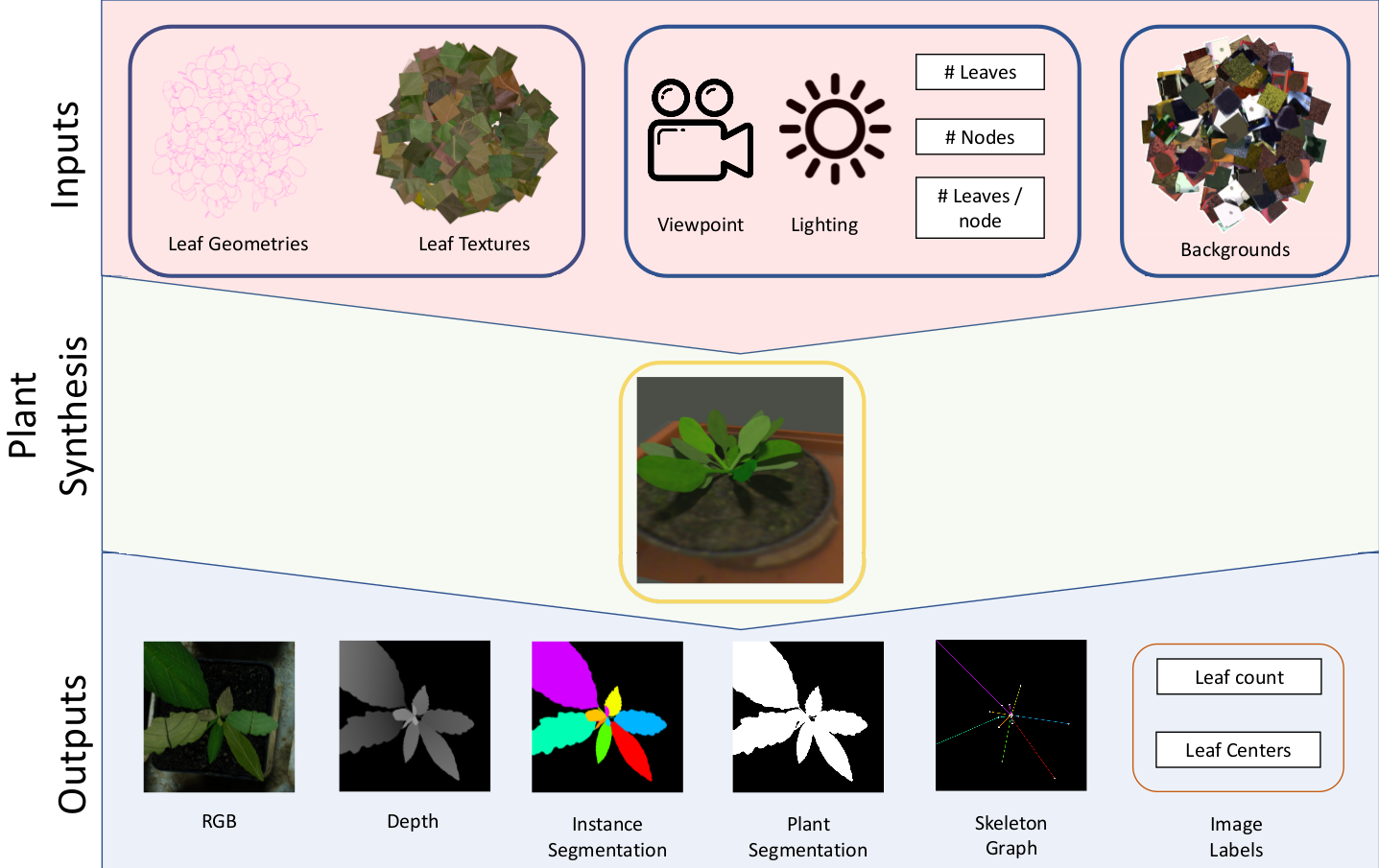}
    \caption{An overview of \textit{UPGen}. A 3D plant model is synthesised using randomly sampled leaf geometries, textures and plant parameters. Generated outputs comprise of a top-down RGB image, depth image, leaf segmentation mask, plant segmentation mask, leaf count, leaf occlusion mask and 3D plant skeleton.}
    \label{fig_synthetic_overview}
\end{figure*}
Supervised deep learning has significantly progressed the state of the art in many applications from image classification~\citep{russakovsky2015, lin2014} to natural language processing~\citep{halevy2009unreasonable}. However, this would not have happened without an abundance of labeled data and computing power~\citep{halevy2009unreasonable, sun2017revisiting}. 
Many publicly available annotated datasets span a wide range of common and generally applicable tasks. Common datasets such as ImageNet~\citep{russakovsky2015}; MSCOCO~\citep{lin2014}; PASCAL VOC~\citep{mottaghi_cvpr14}; the SUN dataset~\citep{xiao2010sun} and Open Images~\citep{OpenImages2} present almost 25 million images for vision tasks such as object detection. Features learned on datasets such as ImageNet have been shown to transfer well to other tasks or datasets. 

Unlike common class objects (e.g. car, table, chair), the \textit{plants} class objects are deformable and inherently stochastic. Two plants of the same genotype (DNA expression) can have an arbitrary number of leaves and leaf placement which can differ in structure and appearance owing to a disparity in phenotype, a plant's observable characteristics.  A plant's phenotype is subject to its genotype, mutations and the environmental conditions. Within the \textit{leaf} class thousands of different plant species exist presenting many different leaf shapes, structures and appearances. 

In recent years, there has been an increasing interest in developing image-based, automated and non-invasive techniques to estimate plant phenotypic information~\citep{pieruschka2019plant, minervini2015image}. Plant phenotyping is a prerequisite for precision breeding and identifying genes associated with important agronomic traits~\citep{zhao2019crop}. Common measurements essential for rapid phenotype discovery and analysis include but are not limited to plant height, leaf shape, leaf count and leaf area index (LAI). Automating the estimation of plants phenotypic traits using current computer vision and machine learning techniques is challenging because of the cost of data capturing (growing and imaging many plants) and cost of manual annotation of the training data. Sufficiently distributed data must be collected and annotated to provide examples of all mutations, genotypes and environmental conditions affect a plant's phenotype and its appearance.

The Computer Vision Problems in Plant Phenotyping (CVPPP) drew the attention of the machine learning community to these challenges with the release of the seminal leaf instance segmentation dataset and challenge in 2014~\citep{scharr2014annotated}. Focusing on rosette plants, the dataset contains two different plant species and multiple mutations within species~\citep{minervini2016finely, scharr2016leaf}. However, many methods do not generalise and still require re-training or modifications to perform consistently across different plant species and mutations~\citep{tsaftaris2019sharing}.

With the costs of growing numerous physical plants, imaging them at one or more growth stages and then annotating the images. It is not tractable to approach \textit{species gap} challenges by collecting more data and training a new model for each plant species, mutation or genotype.

To overcome the many challenges associated with image-based phenotyping, some studies have addressed the lack of annotated plant data through the introduction of synthetic data~\citep{giuffrida2017, ubbens2018use, Kuznichov_2019_CVPR_Workshops, zhu2018data}. %
Current synthetic data solutions, including our previous work~\citep{ward2018deep}, are limited by their ability to generalise to unseen data, the reality gap between the simulator and the real world and the range of  ground truth annotations they support for plants phenotypic traits estimation . %
In our previous work~\citep{ward2018deep} we achieved state-of-the-art results in the CVPPP leaf segmentation challenge (LSC) by using synthetic training data. However, the synthetic images were specific to a single plant species, \textit{Arabidopsis Thaliana}. The previous method also did not consider the high intra-class (leaf) variance and number of parameters required to model plants. The trained model weakly modeled the stochasticity of biological mutation and did not transfer to unseen plant species (\textit{species gap}).

The focus of this work is to inform the use of domain randomised synthetic data in enabling the automation of image-based plant phenotyping. First we present \textit{UPGen}: a Universal Plant Generator for bridging the \textit{species gap}. \textit{UPGen} (Figure~\ref{fig_synthetic_overview}) is a generalised synthetic data pipeline for modeling and generating data of any plant in a top-down high throughput plant phenotyping system.  Our approach leverages Domain Randomisation (DR) concepts to model stochastic biological variation between plants of the same and different species (\textit{species gap}). The same approach also improves model performance across different datasets or imaging environments (\textit{domain gap}).
The main contributions of this paper are as follow:
\begin{itemize}
    \item a novel synthetic data pipeline to generate top down RGB images of plants with leaf instance segmentation masks;
    \item a set of experiments which inform the design, implementation and use of domain randomised synthetic data in automated image based plant phenotyping of any species of plant;
    \item state of the art performance on the CVPPP Leaf Segmentation Challenge;
    \item the largest and most diverse annotated synthetic plant dataset with per-pixel ground truth annotations and pretrained model are made publicly available to accelerate new lines of research in image-based plant phenotypic estimation; and
    \item by presenting a large synthetic multi-modality plant dataset we promote new research avenues which encourage investigation into the use of additional imaging modalities for plant phenotyping without the costs of plant growth, data collection and annotation.
\end{itemize}

%% file: chapters/2_relatedWork.tex
\section{Related Work}\label{sec_relatedWork}

Deep learning segmentation architectures in computer vision are trained using large image datasets with pixelwise annotations. Such datasets include~\cite{Geiger2013IJRR, russakovsky2015, lin2014} and~\cite{scharr2016leaf}. Data collection and annotation is prone to error; time consuming and expensive. Further, data collection may not guarantee a dataset which provides exhaustive training examples. This is particularly relevant to plant phenotyping applications where data collection incorporates the investment of growing individual plants from seed in high-throughput specialised facilities. 
These drawbacks have motivated alternatives when faced with small training data quantities. Namely, the generation of synthetic data and procedures for sample efficient training. Synthetic data is also used to provide image labels in complex and difficult to annotate scenarios such as occlusion~\citep{fulgeri2019can} or being able to synthesise 'hard examples' for improved generalisation~\citep{bozorgtabar2019learn}.
When designing computer vision systems, their ability to generalise to unseen samples is paramount. \textit{Domain gap} refers to the disparity between data, \textit{i.e.} two datasets of similar purpose obtained in different conditions. Further, the \textit{reality gap} describes the gap between synthetic and real data. Domain adaptation is a problem in computer vision and machine learning which addresses the negative impact on generalisation performance due to a distribution mismatch between training and test data. Data augmentation, now standard approach in deep learning applications, involves randomly cropping, flipping and applying photometric variations to training images to combat \textit{domain gap}, reduce overfitting and improve generalisation.
Three types of synthetic data generation methods are described below: cut and paste imaging, generative models and simulated data. %

\textbf{Cut and paste imaging} methods describe the synthesis of new data samples by combining the foregrounds and backgrounds of existing real images~\citep{dwibedi2017cut}. This approach vastly increases the available data and has yielded significant performance increases. Early approaches used random sampling to place foreground objects in background scenes~\citep{dwibedi2017cut}. Such methods can result in unrealistic data which limits generalisation. This was demonstrated by \cite{dvornik2018modeling} where the importance of visual context in foreground placement was demonstrated. Prior work~\citep{dvornik2018modeling, Kuznichov_2019_CVPR_Workshops} have manually encoded contextual cues into their data generation. More recent work has investigated learning such foreground placement and avoiding modeling explicit context~\citep{tripathi2019learning} by combining cut and paste imaging and adversarial learning.

\cite{Kuznichov_2019_CVPR_Workshops} used this technique to construct new images of plants from leaves extracted from a labeled plant dataset. Their approach is effective and achieved state-of-the-art results on the CVPPP competition when generating data from the CVPPP training data. Their method is also evaluated on a dataset of Avocado images, however, to do this a different data construction process designed and implemented. Our method presented here is similar to \cite{Kuznichov_2019_CVPR_Workshops}, however, builds on several identified limitations in data generation and species generalisation. By constructing a 3D plant model and then rendering synthetic data, \textit{UPGen}, is not limited by the data available to 'cut' from. \textit{UPGen} geometrically models plant parameters such as leaf stems, occlusion and employs domain randomisation which enables greater control of the synthesised dataset distribution. Greater control over the data labels is also achieved. For example, to generate a dataset of plants with long stems a single parameter of \textit{UPGen} is tweaked and if required, one could render separate segmentation masks for leaves and stems in the image. Conversely, in cut and paste imaging,  one must collect and label a dataset of plants with long stems to cut from and individually segmenting leaves and stems requires further manual labour.

\textbf{Generative Models} attempt to learn the distribution of a training dataset such that new samples, with some variations, can be generated. Variational autoencoders (VAEs) and generative adversarial networks (GANs) are generative modelling methods most commonly applied to image data. VAEs were built on traditional autoencoders and incorporated a latent variable which enabled the generation of new data rather than simply decoding an encoded vector. VAEs are trained by optimising the lower variational bound of the data log-likelihood. GANs, however, do not learn an explicit probability density function and tend to yield better results on image data than VAEs. Hence, the majority of recent generative modeling for images literature is dominated by GANs.

(GAN) methods involve training two networks, one to generate realistic samples and one to discriminate between real and fake data samples. 
Like cut and paste imaging, these methods require real data samples to produce synthetic data. The real data is used to initially train the generator. GANs have been used to generate new data samples such as~\cite{lin2018st, song2018constructing}. These methods have also been applied to plant images. ARIGAN~\citep{giuffrida2017} produced a new Arabidopsis leaf counting dataset from a GAN trained on the CVPPP dataset. New data samples were generated to an expected, opposed to ground truth, leaf count and did not contain background texture. \cite{bielski2019emergence} proposed a GAN architecture which learned to place foreground objects into background images. Using the generator as the decoder in a VAE, they then trained a segmentation network. This work focused on a single object in each image and, hence, is not directly applicable to leaf instance segmentation. Few GAN architectures are able to produce both data and segmentation annotations~\citep{sixt2018rendergan}. ARIGAN~\citep{giuffrida2017} noted that leaf edges of their generated data were blurred together. Producing synthetic data using GANs has been shown to model geometry and edges poorly~\citep{zhu2017unpaired}.

Other work has focused on inspiring geometry through rendered data and applying a style transfer GAN to bridge the \textit{reality gap} by changing the appearance of the rendered image~\citep{zhu2018data, bousmalis2018using, barth2018improved}. \citep{zhu2018data} applied cut and paste imaging to assemble new combinations of leaf segmentation masks from the CVPPP leaf segmentation dataset. They then synthesised the corresponding plant image using a GAN trained on the same dataset. While effective on the particular dataset, this method used the same GAN to produce both foreground and background textures. Whereas \textit{UPGen} separates the two allowing simple modification of background textures. We show that modification of background textures significantly effects leaf instance segmentation performance on unseen plant datasets.
There is potential for GAN based methods in modeling complex and hard to label phenomena such as plant growth from video data. Recent work by~\cite{spampinato2019adversarial} developed a GAN architecture to learn temporal motion dynamics of objects from video sequences.

\textbf{Simulated Data}
The use of simulation and 3D modeling engines to render photorealistic synthetic images and accompanying annotations is common~\citep{ros2016synthia, ubbens2018use, muller2018sim4cv}. The main limitation of rendered data is the \textit{reality gap}. The \textit{reality gap} refers to domain differences between synthetic and real images and many recent developments in this area, discussed below, focus on sim2real methods to bridge the reality gap.
Compared to real data, a wider distribution of foreground object scale, pose, spatial position and texture can be obtained using these rendering methods because one can specify them during the design stage. However, because of this fine control, synthetic data design can be time consuming and labour intensive. \cite{barth2018data} presented a high fidelity capsicum annuum dataset for robotic fruit picking simulations. The dataset consisted of 10,500 images, however, these were rendered from only 42 plant models. Each plant model was procedurally generated and based on manual physical plant measurements.

Geometrical and textural information are required to produce such 3D models, however, unlike data augmentation, cut \& paste and GAN synthetic data generation methods; real data is not necessarily required to learn from or dissect to produce more samples. The object shapes used by~\citep{khirodkar2019domain} are defined by 3D models and rendered on real image backgrounds. Work by \cite{ubbens2018use} make use of a parametric model of the Arabidopsis plant species~\citep{prusinkiewicz2002art} to generate a leaf count dataset. The same parametric plant model has also been used to learn latent representations of phenotypic plant treatment responses from synthetic data~\cite{ubbens2019latent}. Because the plant model parameters must be defined, these methods produce data which are species and application specific.

Domain randomisation (DR) reduces the reality gap for synthetic data by applying random augmentation techniques to individual components of a synthetic scene. DR works by sufficiently randomising each synthetic image such that the real world appears as another permutation in the randomness. Training on data rendered with random texture, lighting, camera position and with distractor objects results in a system invariant to domain variations. Domain randomised synthetic data can be significantly more widely distributed than real data. For example, \cite{tremblay2018} showed this by comparing the distributions of cars per image in the KITTI dataset~\citep{Geiger2013IJRR} to their domain randomised synthetic data. DR was initially used to train a reinforcement learning system for a grasping task on a basic simulation and transfer it to the real world~\citep{tobin2017domain}.

The idea of DR has been applied to bridging \textit{reality gap} in many tasks from car detection~\citep{tremblay2018} to detecting packaged food in refrigerators~\citep{rajpura2017transfer}. Specific to image-based plant phenotyping, \cite{ward2018deep} applied DR to leaf instance segmentation. Their synthetic data was specific to a single plant species and did not model leaf stem or plant height growth. When used to train a instace segmentation algorithm, their method achieved state-of-the-art performance on the CVPPP-A1 test dataset. In this work we build on \cite{ward2018deep} by modeling leaf stems, plant height growth and applying DR to overcome both \textit{reality gap} and \textit{species gap} simultaneously. Our presented improvements achieve state-of-the-art performance across all CVPPP test datasets which cover multiple plant species. We also demonstrate its generalisation on unseen plant species and imaging scenarios. 

Training models with limited or no supervision is a research area closely coupled with synthetic data. Initially training on synthetic data and then fine tuning on real, in-domain data is a common method for combined training~\citep{tremblay2018, giuffrida2017}. Other work has explored training a feature extractor on real data, freezing the network weights and training the remaining layers on synthetic data~\citep{hinterstoisser2017}. Further, the simultaneous use of both real and synthetic data can be sorted into methods randomly selecting images for each mini-batch and those ensuring a number of real-synthetic ratio of images in each mini-batch~\citep{ward2018deep}.
Each method has encountered reality gap challenges when applied to a single imaging environment and images of a single plant species. 
In this study, we employ domain randomised rendered synthetic data for image-based plant phenotyping. We validate our method on the challenging task of leaf instance segmentation to show it effectively bridges the reality gap and achieves comparable performance to training on real data. It also bridges the \textit{species gap} with performance generalising to unseen plant species and imaging scenarios. %

%% file: chapters/3_method.tex
\section{UPGen: The Universal Plant Generator}\label{sec_method_synthesiser}
We present our synthetic data generation pipeline as a viable alternative to collecting and annotating real data in top-down image-based plant phenotyping systems. Our data pipeline can be used as a replacement for real data or in conjunction with few real data samples to train deep learning models for plant-level predictions 
(projected LAI, height, plant architecture) or leaf-level predictions (individual leaf area, leaf count, leaf growth rate). 
Further, this trained model is shown to overcome \textit{species gap} and \textit{domain gap} better than a model trained on only real data. Figure~\ref{fig_synthetic_overview} displays an overview of our proposed method. Our pretrained model and a copy of our dataset is available at \url{https://csiro-robotics.github.io/UPGen_Webpage/}.

Our proposed \textit{UPGen} pipeline is based on the assumptions that (i) a single plant is visible in each image; (ii) a plant can be considered as an arrangement of individual leaves in 3D space; and (iii) in top-down imaging it is sufficient to treat a single leaf as planar. 

First leaf geometries and textures are sampled and processed from publicly available leaf datasets. Next a set of background textures are collected from online datasets and by inpainting plant images. Finally, the geometries and textures are used to assemble a synthetic plant and render a synthetic data sample.

Combined with a set of specified pipeline parameters a synthetic dataset can be generated. Domain randomisation is employed at each step of the pipeline to produce a diverse and generally applicable dataset. Each output sample consists of; an top-down RGB image, a corresponding depth map, leaf instance segmentation mask, plant segmentation mask, leaf count, leaf occlusion masks and 3D plant skeleton graph.
Each step of the process is further explained in the following sections. 
Specific technical details regarding the implementation of the proposed pipeline can be found in supplementary material~\ref{appendix_implementation_details}.

\begin{figure}[ht]
	\centering
	\subfigure[3D Plant Model]{
		\includegraphics[width=0.45\textwidth]{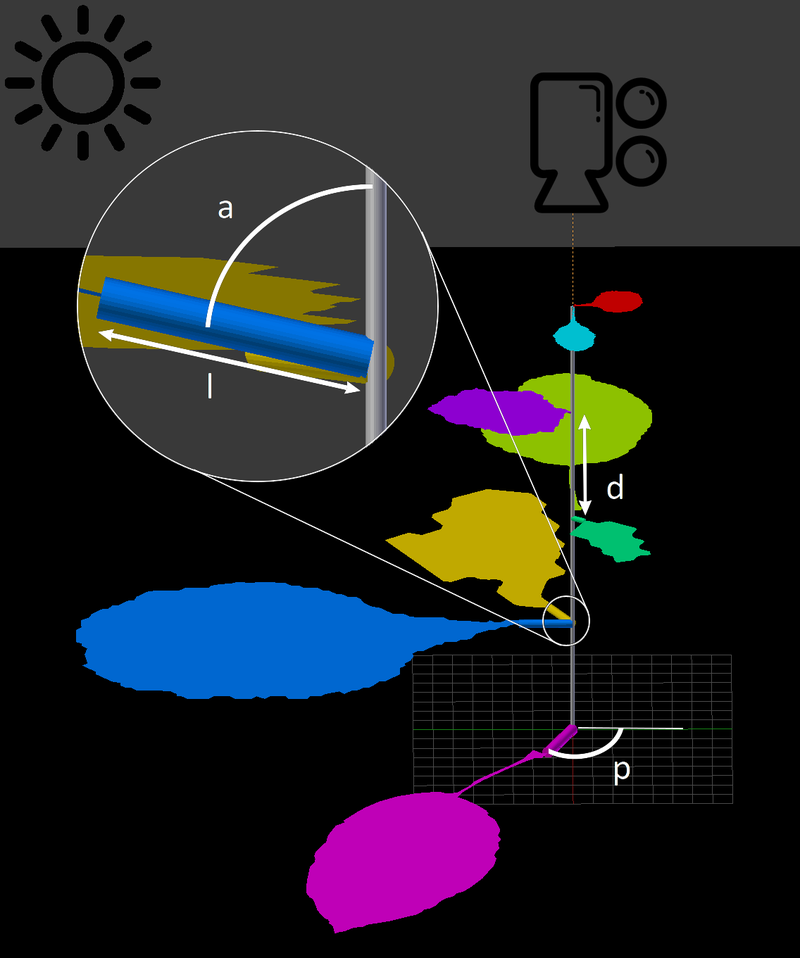}
	}
	\hspace{-5mm}
	\subfigure[RGB]{
		\includegraphics[width=0.2\textwidth]{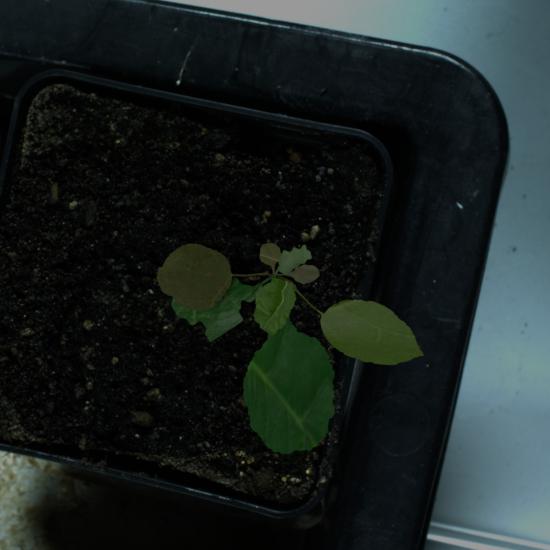}
	}
	\hspace{5mm}
	\subfigure[Segmentation Mask]{
		\includegraphics[width=0.2\textwidth]{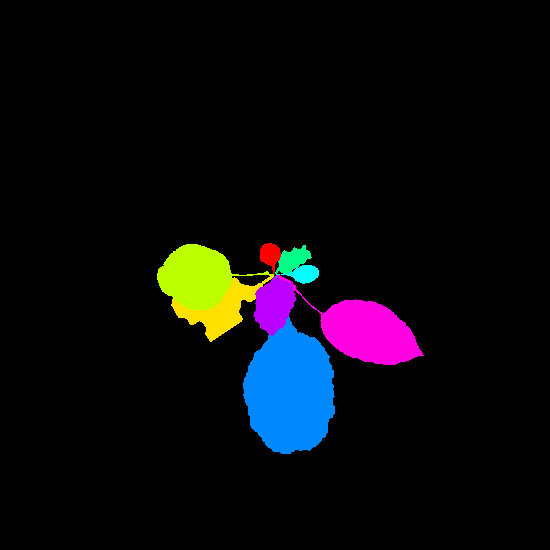}
	}
	\caption{A visualisation of a 3D plant model (a) and the corresponding rendered RGB image (b) and leaf instance segmentation mask (c). This particular sample has $N=8$ leaves. The stem angle ($a$); stem length ($l$), phyllotaxy angle ($p$) and leaf node separation distance ($d$) are labelled on the model in white.}
    \label{fig_plant_assembly_params}
\end{figure}

\subsection{Leaf Geometry Processing}\label{sec_method_leaf_geometry}
A database of 17,957 inspiration leaf geometries from 46 different plant species was collected from existing real data sets~\citep{scharr2016leaf, lee2015deep, migicovsky2018morphometrics}. Geometry processing involved identifying the leaf contour, vertically aligning it to a canonical orientation and then applying the Delaunay triangulation algorithm to describe the shape as a planar mesh. The geometry processing methods differed subtly for each dataset, details of these are described in supplementary material~\ref{appendix_leaf_geometry}. The set of leaf geometries are shown in the top left of Figure~\ref{fig_synthetic_overview}.

\subsection{Leaf Texture Processing}\label{sec_method_leaf_texture}
A database of 40,552 leaf textures from multiple sources to provide a wide range of textural information was collected from existing real data. Namely the low frequency content textures from~\citep{ward2018deep}; high frequency content textures (containing strong edges such as leaf veins) from open licence texture databases and texture patches from the leaves of the MalayaKew dataset~\citep{lee2015deep}. Texture patches were rectangular image patches extracted from within the border of a leaf. To add further variance to the set of leaf textures, augmentations were applied to the images. Details of specific augmentations are outlined in supplementary material~\ref{appendix_leaf_texture}.

\subsection{Background Texture Processing}\label{sec_method_background_texture}
In image-based plant phenotyping platforms, such as~\citep{scharr2016leaf} and~\citep{moghadam2017plant}, the background remains fairly consistent. In context backgrounds were obtained from existing images by inpainting the visible plant. This procedure is also used by~\citep{ward2018deep} and~\citep{Kuznichov_2019_CVPR_Workshops}.
In our experiments, we generated synthetic data using 18 background textures consisting of imaged soil and inpainted images from the CVPPP datasets. Similar to the leaf textures, image augmentations were applied to the background textures to add further variance. Details of specific augmentations and the inpainting procedure are outlined in supplementary material~\ref{appendix_background_texture}.

\subsection{Plant Assembly}
\label{sec_method_plant_assembly}
 To generate a synthetic training sample, a 3D plant model is assembled and rendered on a background with a random selection of leaf textures. The leaves are arranged in nodes equally spaced over the plant height. The number of leaves ($N$), plant height and number of leaf nodes are sampled for each generated plant. $N$ leaves are randomly selected from the pool of inspiration leaves and shape permutations are applied to each by randomly scaling along each axis. Each leaf is assigned a node and scaled accordingly to emulate a natural growth pattern of mature leaves being larger and existing lower on the plant stem. At each node, uniform leaf phyllotaxis (location around the stem) is achieved by sampling a uniformly distributed random variable for each leaf. Described in polar coordinates, $z$ is set to the node height and angle $(0, 2\pi)$ is sampled from the distribution. The phyllotaxis angle, $p$, is visualised on a 3D plant model in Figure~\ref{fig_plant_assembly_params}. The diversity of plant leaf arrangements were modeled using normally distributed random variables. These parameters were the roll; pitch; yaw; stem length (distance from the plant trunk, $l$) and axil (stem angle, $a$). The parameters of their distributions were chosen to reflect natural plant characteristics. For example (the faces of the leafs will normally be facing up towards the light source). 
 
 By randomising each of the previously mentioned plant parameters; capturing a range of leaf shapes from multiple leaf data sources and randomising leaf and background textures a synthetic dataset of plant images can be generated. Figure~\ref{fig_syntheticVisual} visualises example images and their corresponding leaf instance segmentation masks. Our synthetic data models a number of plant species and plant of various growth stages. 

\subsection{UPGen Outputs}
\label{sec_method_outputs}
The outputs of the \textit{UPGen} pipeline are visualised in Figure~\ref{fig_synthetic_overview}. For each synthetic plant the following outputs are produced; an RGB render, a depth map visualisation, a leaf instance segmentation mask, a plant segmentation mask, the plant skeleton graph, leaf count and locations of leaf centers. The depth map is produced from distance along the $z$ axis to each object from the ray tracing render. The skeleton graph describes the key points (leaf stem, petiole and leaf tip) and connection between them for each leaf in 3D space.

%% file: chapters/4-1_Exp.tex
\section{Experimental Design}
\label{sec_method_experiments}

\begin{figure}[thpb!]
	\centering
	\subfigure{
		\includegraphics[width=0.12\textwidth]{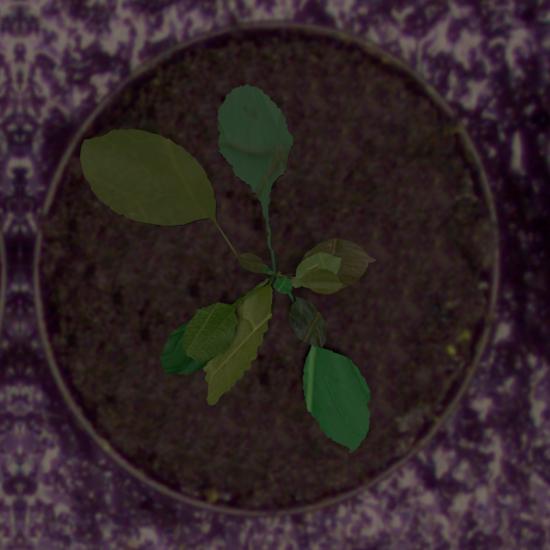}
	}
	\hspace{-5mm}
	\subfigure{
		\includegraphics[width=0.12\textwidth]{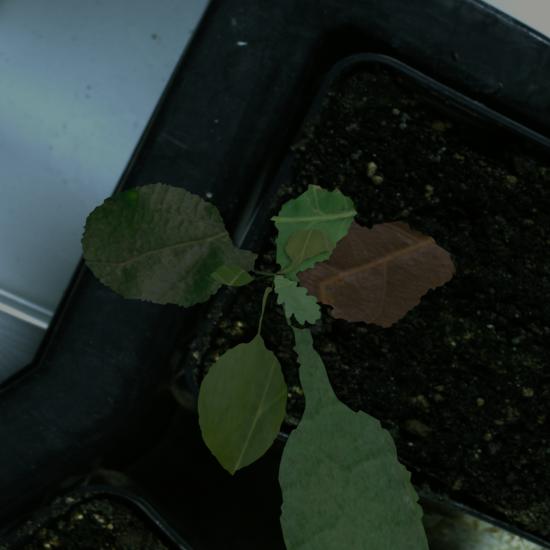}
	}
	\hspace{-5mm}
	\subfigure{
		\includegraphics[width=0.12\textwidth]{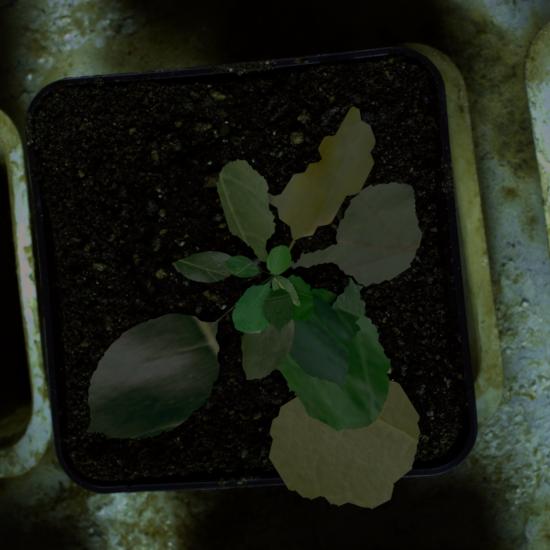}
	}
	\hspace{-5mm}
	\subfigure{
		\includegraphics[width=0.12\textwidth]{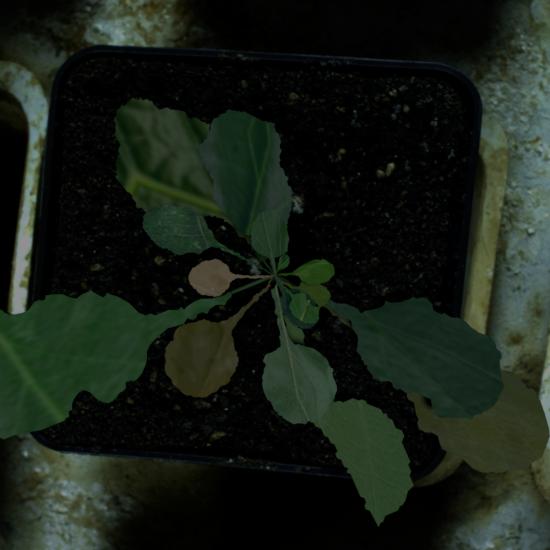}
	}
	
	\vspace{-7.5mm}
	\hspace{-6mm}
	\newline
	\subfigure{
		\includegraphics[width=0.12\textwidth]{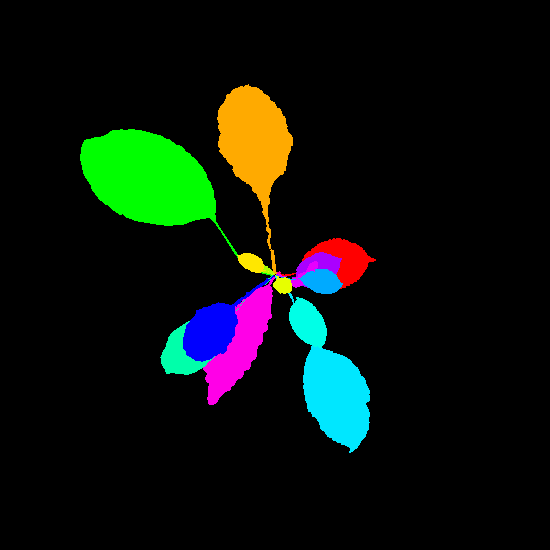}
	}
	\hspace{-5mm}
	\subfigure{
		\includegraphics[width=0.12\textwidth]{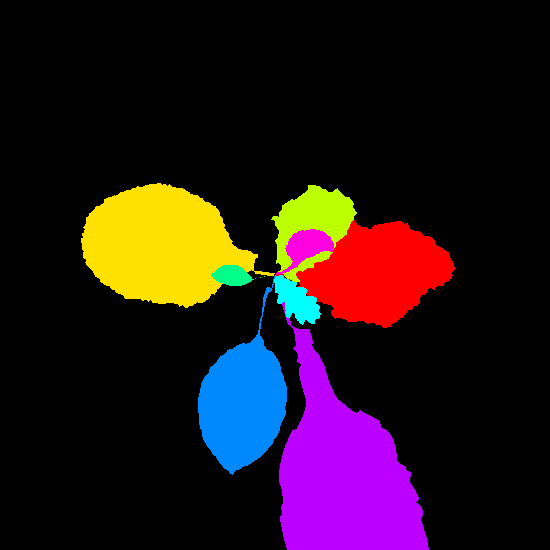}
	}
	\hspace{-5mm}
	\subfigure{
		\includegraphics[width=0.12\textwidth]{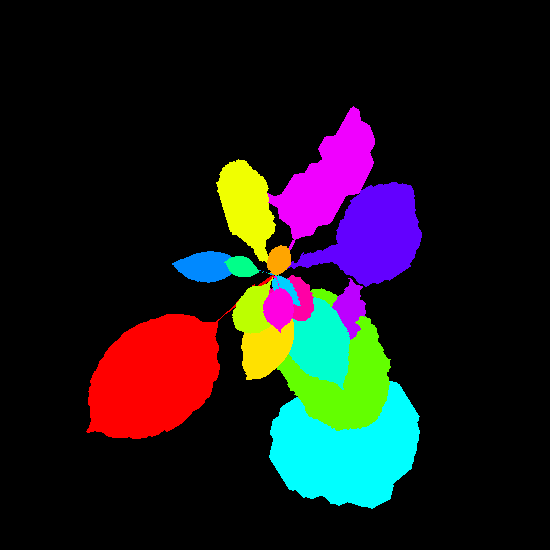}
	}
	\hspace{-5mm}
	\subfigure{
		\includegraphics[width=0.12\textwidth]{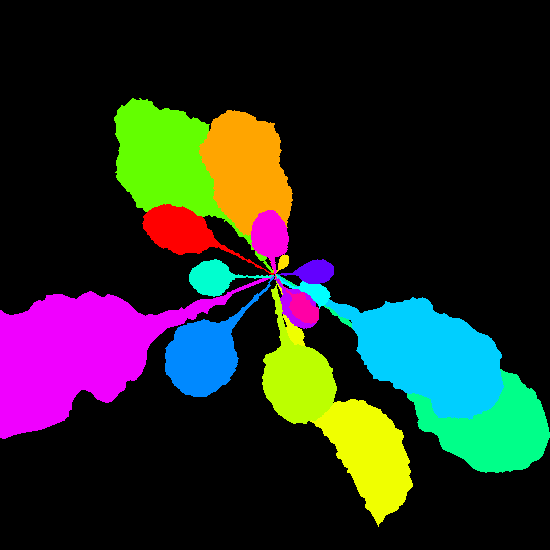}
	}
	\caption{Example images and corresponding ground truth leaf instance segmentation masks from the synthetic dataset.}
	\label{fig_syntheticVisual}
\end{figure}

\textit{UPGen} produces training data for image-based plant phenotyping applications without the cost of data collection or manual annotation. We validate it by applying it the task of leaf instance segmentation. It is an important task in image-based plant phenotyping because it enables plant measurements such as plant segmentation and leaf counting. Instance segmentation is also important because of the high cost to obtain accurate per-pixel ground truth annotations.

\subsection{Leaf Instance Segmentation Architecture}\label{sec_method_leaf_segmentation}
We benchmark the use of \textit{UPGen} by training a leaf instance segmentation model based on the Mask-RCNN architecture~\citep{he2017mask}. This achieved state-of-the-art results in the 2017 COCO instance segmentation task~\citep{lin2014} and is used by the current CVPPP leaf segmentation challenge state of the art~\citep{ward2018deep}. The model consists of a feature extractor and a Region Proposal Network (RPN). The network heads produce bounding box classification; bounding box regression and an object mask from each proposed region. We used the ResNet101 backbone with a feature pyramid network for Mask-RCNN. Like~\citep{ward2018deep}, 256 regions of interest per image were used for training. Each experiment involved splitting the training data into 80\% training and 20\% cross validation. 
Models were trained using stochastic gradient descent optimiser with a learning rate of 0.001 and momentum and weight decay terms of 0.9 and 0.001 respectively. The batch size was fixed to four. The same parameters were also used for experiments involving fine-tuning.

Except in the ablation study, all training procedures utilised standard augmentation techniques. Namely: vertical flips; horizontal flips; cropping or padding the image by up to 25\%; applying Gaussian blurring; random scaling between 80\% and 120\%; random rotation, translation and shearing; per channel image brightness changes and, lastly, random changes to hue and saturation between -20 and 20. Each augmentation was applied in a random order and with an independent 50\% chance of being applied to each input image. During training, the model checkpoint with the lowest cross validation loss was used. The results of our leaf instance segmentation experiments are present using the Symmetric Dice (SBD) segmentation metric. Formulation of SBD can be found in supplementary material~\ref{appendix_dice_definition}.

\subsection{UPGen Synthetic Training Data}\label{sec_method_training_data}
 For all synthetic experiments in this paper, 10,000 samples were generated using the \textit{UPGen} pipeline. Using the parameters described in section~\ref{sec_method_synthesiser} images of resolution 550 x 550 pixels were used for training. Figure~\ref{fig_syntheticVisual} shows example images from the \textit{UPGen} pipeline.

\subsection{The Computer Vision Problems in Plant Phenotyping Data}\label{subsec_cvpppData}
\begin{figure}[t!]
	\centering
	\subfigure{
		\includegraphics[width=0.18\textwidth, height=0.18\textwidth]{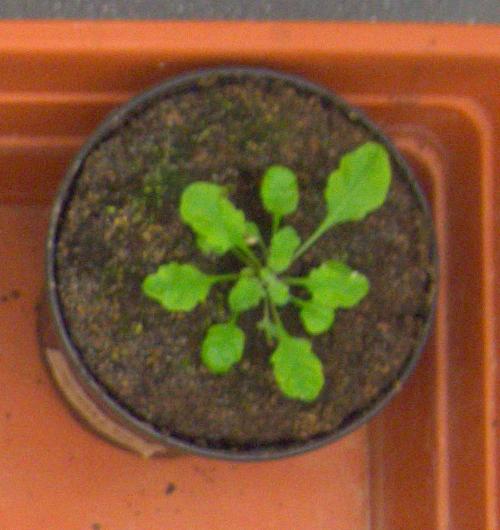}
	}
	\hspace{-5mm}
	\subfigure{
		\includegraphics[width=0.18\textwidth, height=0.18\textwidth]{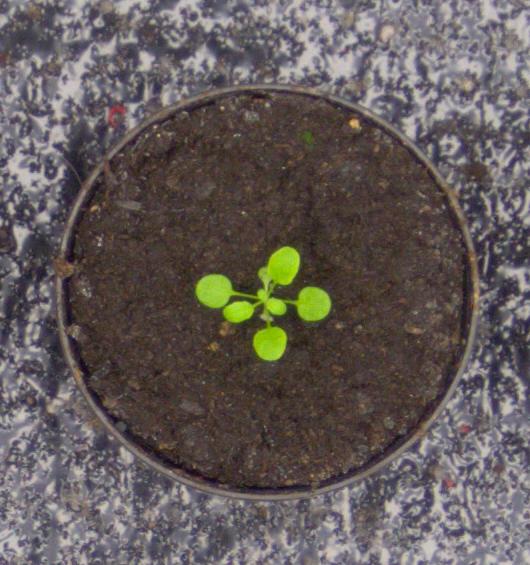}
	}
	\hspace{-5mm}
	\subfigure{
		\includegraphics[width=0.18\textwidth, height=0.18\textwidth]{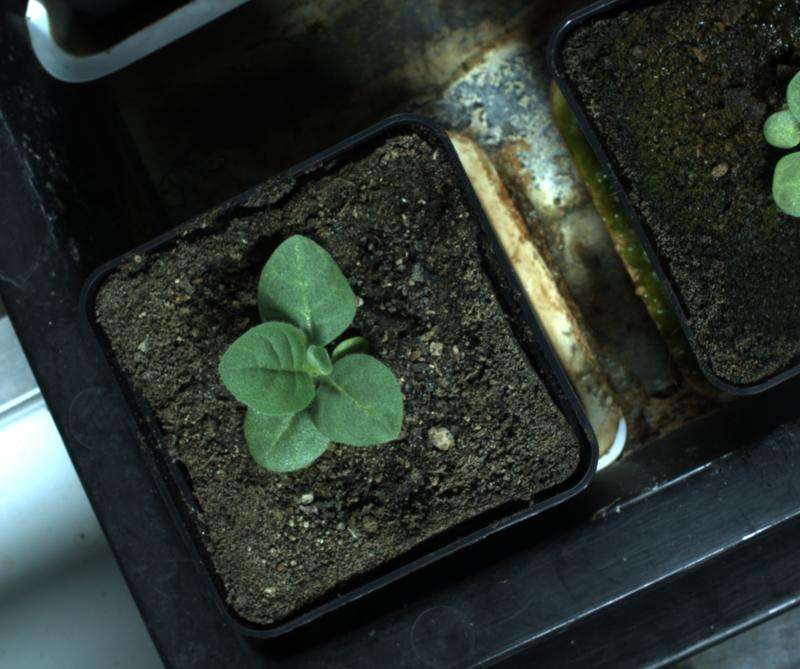}
	}
	\hspace{-5mm}
	\subfigure{
		\includegraphics[width=0.18\textwidth, height=0.18\textwidth]{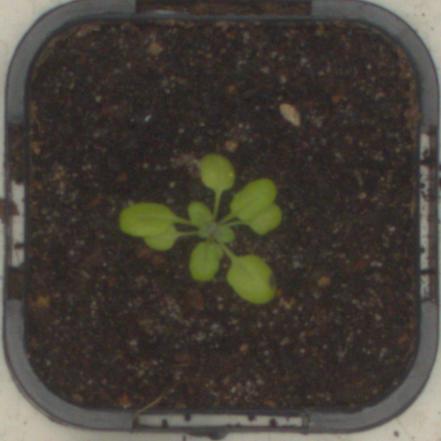}
	}
	\caption{Selected images from the real datasets (CVPPP A1 - A4 are at the top left, top right, bottom left and bottom right respectively) used in this study~\citep{minervini2016finely}. Note the significant variation in leaf textures, shape and arrangement and the differences in background textures between imaging scenarios.}
	\label{fig_realDataVisual}
	\vspace{-5mm}
\end{figure}

In order to compare our methods with existing literature, our models are evaluated on the Computer Vision Problems in Plant Phenotyping (CVPPP) Leaf Segmentation Challenge (LSC) (2017) dataset~\citep{scharr2016leaf, minervini2016finely, bellAndDeeDataset}. This dataset consists of 4 different sub-datasets, A1, A2, A3 and A4. Each sub-dataset contains images of a different species. We also define \textit{CVPPP-All} as the union of all sub-datasets. It is used as the baseline real data training dataset in this study. Table~\ref{tab_cvppp_dataset_breakdown} breaks down the contents of the CVPPP dataset of top down images of rosette plants. In Figure~\ref{fig_realDataVisual}, the different plants are visualised. The CVPPP test dataset labels are not available to the public. Hence, all results were obtained by evaluating the predictions on the competition CodaLab evaluation server\footnote{\url{https://competitions.codalab.org/competitions/18405}}.

\begin{table}[t!]
	\caption{The image size and number of samples in each of the CVPPP LSC datasets. Plant phenotyping datasets of sufficient size for deep learning applications are uncommon owing to the cost of data collection and annotation.}
	\label{tab_cvppp_dataset_breakdown}
	\vspace{2mm}
	\centering
	\begin{tabular}{|p{0.75cm}|p{1.6cm}|p{1.4cm}|p{1.1cm}|p{1.7cm}|}
		\hline
		\textbf{Data} & 
		\textbf{Image\newline Resolution\newline (pixels)} &
		\textbf{Training\newline Images} &
		\textbf{Test Images} & 
		\textbf{Plant \newline Species} \\
		
		\hline
		A1 &
		500 x 530 &
		128 &
		33 &
		Arabidopsis Thaliana\\
		
		\hline
		A2 &
		530 x 565 &
		31 &
		9 &
		Arabidopsis Thaliana\\
		
		\hline
		A3 &
		2448 x 2048 &
		27 &
		65 &
		Tobacco \newline (N. tabacum) \\
		
		\hline
		A4 &
		441 x 441 &
		624 &
		168 &
		Arabidopsis Thaliana\\
		\hline
	\end{tabular}
\end{table}

\subsection{In The Wild Test Data}
\begin{figure}[b!]
	\centering
	\hspace{-5mm}
	\subfigure{
		\includegraphics[width=0.18\textwidth, height=0.18\textwidth]{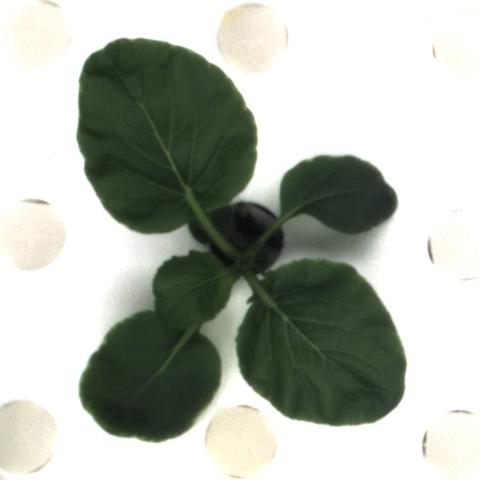}
	}
	\hspace{-5mm}
	\subfigure{
		\includegraphics[width=0.18\textwidth, height=0.18\textwidth]{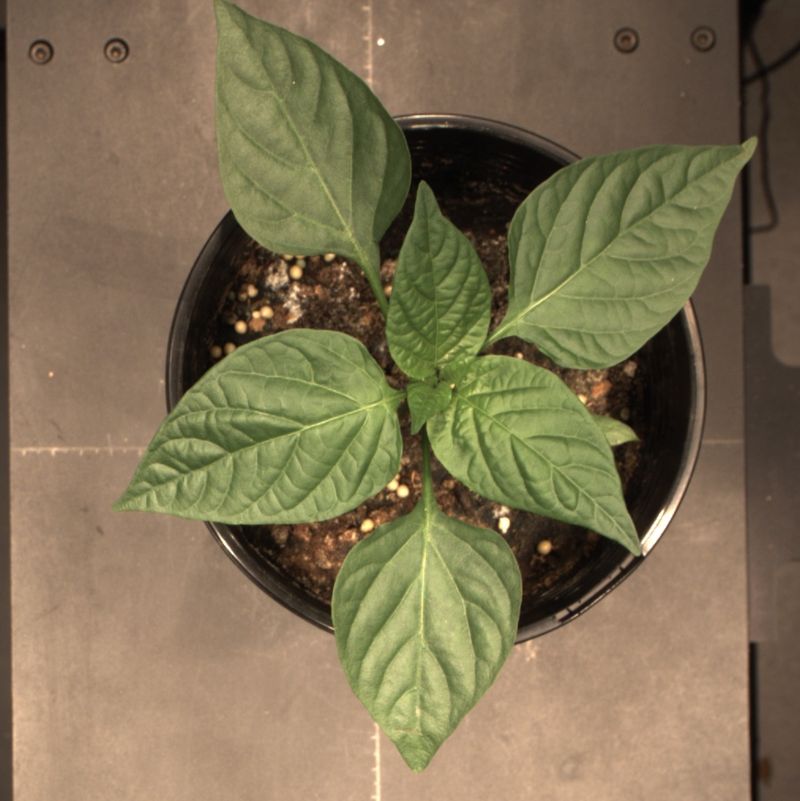}
	}
	\caption{Example images from the \textit{in the wild} datasets used in this study. A Komatsuna~\citep{uchiyama2017easy} (left) and capsicum (right) plant are shown. Note the differences between leaf shape, appearance and geometry between leaf species and the differences in imaging environments to the real data used for training in Figure~\ref{fig_realDataVisual}. The capsicum image was cropped for visualisation.}
	\label{fig_unseenDataVisual}
\end{figure}

To demonstrate the ability of an \textit{UPGen} trained model to generalise to completely unseen data, our models are evaluated on two unseen datasets of different plant species and imaging systems. These consisted of images from the Komatsuna~\citep{uchiyama2017easy} dataset and a proprietary capsicum dataset which are visualised in Figure~\ref{fig_unseenDataVisual}. At no point were these images directly trained on. We show performance improvements when training on synthetic data with in-context background textures from these datasets. Foreground (plant/leaf) geometry and texture information, however, remains unseen.

We present results on a test dataset of 225 images from the Komatsuna~\citep{uchiyama2017easy} dataset. The full dataset was collected by imaging 4 plants on an hourly basis. To avoid misleading results on many images which are visually similar, we used every $4^{th}$ image in the dataset. The dataset we present results on contained the full range of plant ages and sizes present in the full dataset.
Komatsuna plants are classified as rossette similar to the CVPPP dataset. To evaluate a greater \textit{species gap}, we present results on capsicum plants. This species, \textit{capsicum annuum}, is not a rossette and exhibits vertical growth. Further, the leaf shape and arrangement differs greatly from the Arabidopsis in the CVPPP dataset. To collect this dataset, 20 capsicum seedlings were grown in a greenhouse and imaged, top-down, at different growth stages. We manually annotated ground truth segmentation labels for 20 images, the image resolution is 2736 x 2192.

%% file: chapters/4_results.tex
\section{Results}\label{sec_results}

In order to validate \textit{UPGen}, we evaluate it on the task of leaf instance segmentation. This task in image-based plant phenotyping was chosen because it is most data hungry task and enables several common phenotyping traits measurements; leaf count, plant area, LAI, leaf center locations and leaf morphology. 
We present the results of several experiments exploring the use of synthetic data in an image-based plant phenotyping setting. Specifically \textit{UPGen} is validated on the CVPPP Leaf Segmentation Challenge (LSC) and on a \textit{species gap} problem to demonstrate its applicability in high throughput imaging of unseen plant species datasets \textit{in the wild}. On these datasets, we compare our approach to the state-of-the-art and a baseline method, training on the small amount of real data available. We also investigate two different methods of combining real and synthetic data for training, batch balanced training and fine tuning. These are common techniques to improve \textit{domain gap} and boost the performance on real test datasets. Finally, an ablation study is conducted to reveal the effects dataset size and of each component of the \textit{UPGen} pipeline on leaf segmentation performance.

\subsection{Using Synthetic data}\label{sec_results_improvement_real}

\input{tables/4_batchBalance.tex}

We explore the utility of \textit{UPGen} synthetic data by using it to train a model and demonstrate it's performance on the CVPPP LSC. We compare to a baseline, CVPPP-All, which is the same model trained on the CVPPP competition training data as defined in Section~\ref{subsec_cvpppData}.
\vspace{-4mm}

\subsubsection{Replacing Real Data}
To demonstrate that \textit{UPGen} data is realistic and a valid alternative to real data we train a model on only synthetic data. Evaluating it on the CVPPP LSC, results for the four test datasets and mean performance are presented in Table~\ref{tab_results_balanced_training}. Compared to the baseline, CVPPP-All, the \textit{UPGen} trained model achieved higher performance on sub-dataset A1 and equal performance on A2 and A4. On A3 performance was 1.2\% lower, however, a higher average result across the four datasets was achieved. This suggests that the synthetic data is realistic and a valid alternative to collecting and annotating real data. The wide range of \textit{UPGen} plant images does not appear to exhibit domain gap challenges given that comparable performance is achieved and the training and test images of the CVPPP LSC are from the same plant species and imaging environment.

\subsubsection{Combining with Real Data}
We further explore the use of \textit{UPGen} data by training a model on a combination of real and synthetic data. Fine tuning and batch balanced training were two methods of combining datasets that were investigated. These methods improve performance by reducing the \textit{reality gap} between training and test data. For each combination technique, the trained model was evaulated on the CVPPP LSC and compared to baseline methods: training on real data only (CVPPP-All) and training on synthetic data only (\textit{UPGen}). Results are presented in Table~\ref{tab_results_balanced_training}.

\textbf{Batched Balanced Training.} In batch balanced training, each mini-batch contains equal numbers of synthetic and real (CVPPP-All) data samples. This offers a form of regularisation where gradient update steps are hypothesised to be in favour of the real data and overfitting is discouraged by the randomised and widely distributed synthetic data. Using batch balancing, significantly improved leaf instance segmentation performance is achieved over the baselines for A1, A2 and A4 sub-datasets resulting in greater average performance. Reduced performance occurred on A3 because that plant species is under represented in the real data used for training, CVPPP-All (see sub-dataset representations in Table~\ref{tab_cvppp_dataset_breakdown}). The improved performance here suggests that batch balanced training improves performance when real data of a relevant plant species is available.

\textbf{Fine Tuning.} Fine tuning is a commonly employed domain adaptation technique across many deep learning computer vision applications which involves initialising a model on weights trained on a large general dataset, such as ImageNet, and then further fine tuning the weights on data from the particular application or domain. The intuition behind this approach is that the features learned on such general datasets are applicable to new tasks. No comparison to training on large general dataset was made because large datasets used for these purposes, such as ImageNet or MS COCO, contain few to no plant images and tend to be focused on a very different task, object detection. The results of fine tuning are compared to real data and synthetic data baselines in Table~\ref{tab_results_balanced_training}. An increase in performance was achieved using fine tuning for all sub-datasets of the CVPPP LSC. Achieving greater performance improvements than balanced batch training, this result suggests that fine tuning is a preferable real data combination method.

\begin{figure}[t!]
    \vspace{-3mm}
  \centering
    \subfigure[A1]{
        \includegraphics[width=0.22\textwidth]{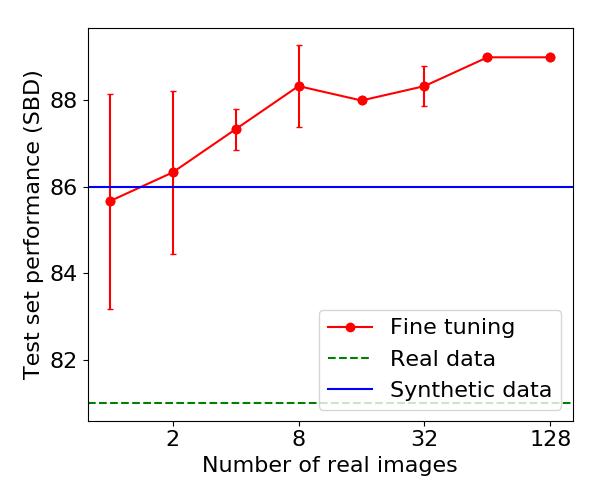}
    }
    \hspace{-3mm}
  	\subfigure[A2]{
        \includegraphics[width=0.22\textwidth]{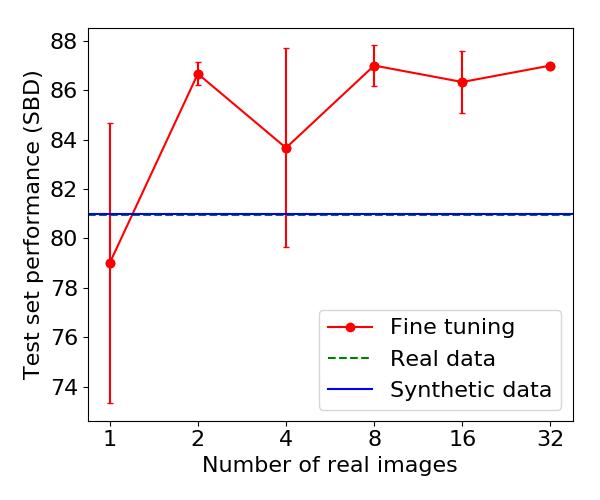}
    }
    \newline
    \subfigure[A3]{
        \includegraphics[width=0.22\textwidth]{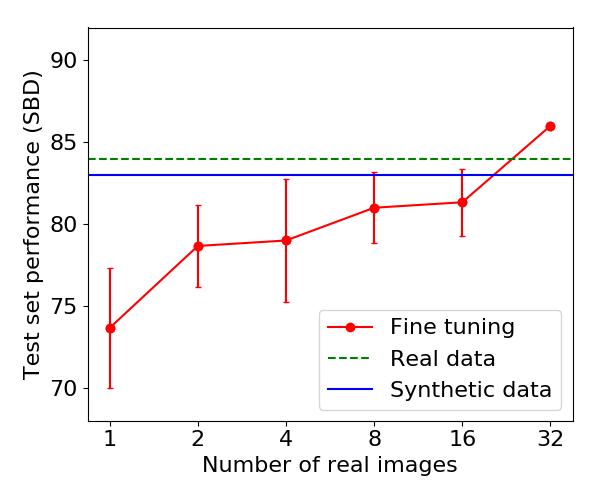}
    }
    \hspace{-3mm}
  	\subfigure[A4]{
        \includegraphics[width=0.22\textwidth]{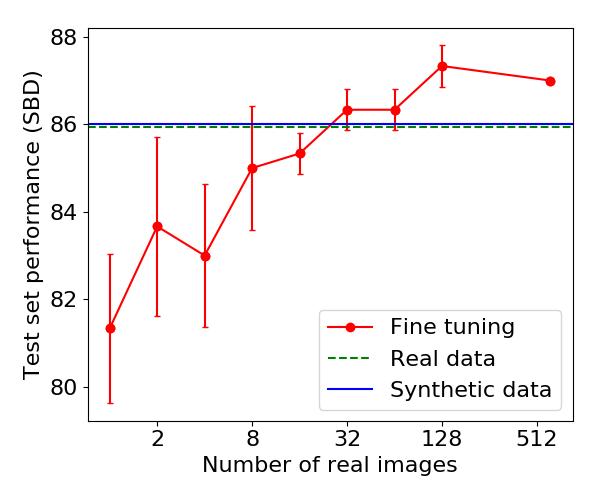}
    }
    \vspace{-3mm}
    \caption{Model performance when fine tuning on different numbers of images compared to baselines: training on synthetic (solid horizontal line) and training on real data only (CVPPP data subsets)(dashed horizontal line). Each data point represents the mean of three replicates. Note how the standard deviation of the data points tends to decrease as more real images are used. This is expected as the model performance is very sensitive to the random choice of a single/few real training images.}
    \label{fig_finetune_data_amount}
    \vspace{-1mm}
\end{figure}

\textbf{Species Specific Fine Tuning.} To further explore the benefits of fine tuning, we investigate how the performance improvements change with the number of real data samples used. This experiment may be informative when budgeting for data collection in an automated image-based plant phenotyping setting. For each sub-dataset in the CVPPP dataset, we take a model, pretrained on \textit{UPGen} and fine tune it on different amounts of real data samples from that sub-dataset. Performance is evaluated on the corresponding test sub-dataset from the CVPPP LSC. When fine tuning on small amounts of real data, results could be effected by the choice of specific data samples. To account for this, each result we present is the mean and standard deviation of three replicates where a different selection of training data samples is ensured. Figure~\ref{fig_finetune_data_amount} presents the results for each sub-dataset. In each plot, the horizontal blue and dashed green lines represent the baselines, a model trained on \textit{UPGen} data and real data (CVPPP-All) respectively. On all plots a trend of decreasing standard deviation with more real images used for fine tuning is seen. This shows that as more images are sampled, the performance is less reliant on the choice of a single/few real images. Across all sub-datasets, fine tuning outperformed the baselines. For A1 and A2, this required as few as two training samples. While 32 training samples were required on A3 and A4. These results suggest that the number of images required for fine tuning depends on the size of the \textit{domain gap} between the synthetic data used for pretraining and desired application dataset. Sub-datasets, A1 and A2, are similar plant species and required a similar amount of fine tuning data. Unlike the synthetic data used for pretraining, the tobacco plants in A3 do not have any visible plant stems which increased the \textit{domain gap} between them. It is expected that adjusting the \textit{UPGen} stem length parameter could reduce the gap. Similarly, the A4 dataset contains a wide range of plants from different growth stages. The strength of \textit{UPGen} here is the ability to reduce the \textit{domain gap} by adjusting the required parameters, produce additional training data for free and improving performance.

Fine tuning is preferable to the batch balanced method when considering generalisation. Batch balancing can be applied when a significant number of labeled training samples are available for training from scratch. Fine tuning, however, achieves comparable or better performance by requiring significantly less labelled training data.

\subsection{Comparison to State-of-the-art}\label{sec_results_comparison_stateOfArt}
\begin{figure}[b!]
	\centering
	\subfigure{
		\includegraphics[width=0.22\textwidth, height=0.22\textwidth]{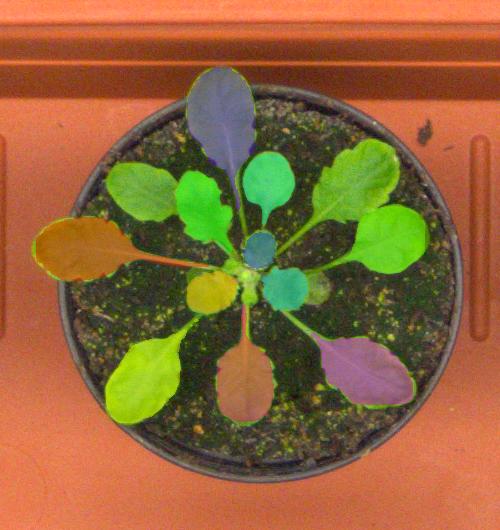}
	}
	\hspace{-5mm}
	\subfigure{
		\includegraphics[width=0.22\textwidth, height=0.22\textwidth]{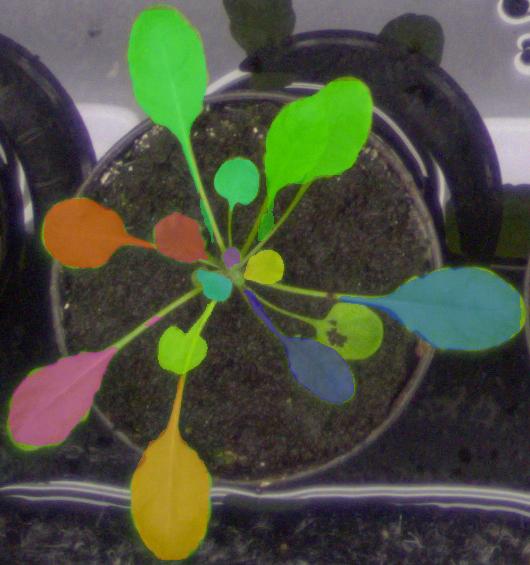}
	}
	\hspace{-5mm}
	\subfigure{
		\includegraphics[width=0.22\textwidth, height=0.22\textwidth]{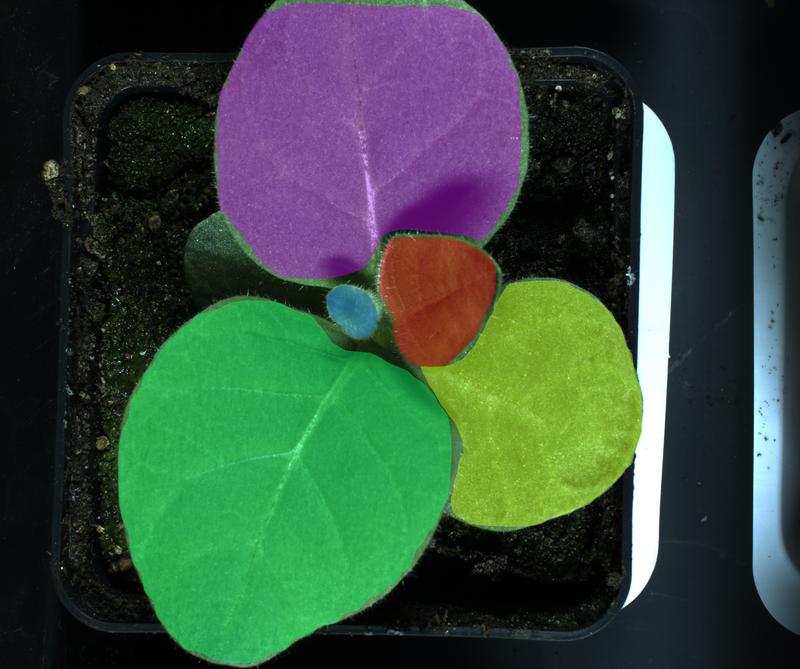}
	}
	\hspace{-5mm}
	\subfigure{
		\includegraphics[width=0.22\textwidth, height=0.22\textwidth]{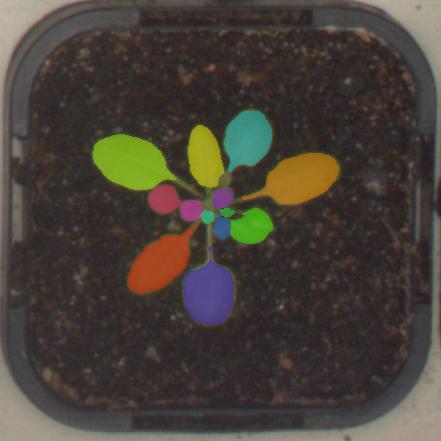}
	}
	\caption{A qualitative assessment of predicted leaf instance segmentation on CVPPP test images (CVPPP A1 - A4 are at the top left, top right, bottom left and bottom right respectively). Predictions made by a model trained only on synthetic data.}
	\label{fig_syntheticResults}
\end{figure}

We compare our best performing models to the state-of-the-art in the public CVPPP LSC. Our results  outperform the state-of-the-art across all but one test datasets and achieves the best mean performance (Table~\ref{tab_results_literature}). \textit{UPGen} performs the most consistently across all datasets as shown by the low standard deviation. These were obtained by training on a combination of real and synthetic data. The result for A1 was achieved employing batch balanced training while those for A2 and A3 utilised fine tuning on the real data. Training on synthetic data alone also yielded competitive results (Table~\ref{tab_results_balanced_training}), outperforming the state-of-the-art on A2 and A3. Figure~\ref{fig_syntheticResults} presents qualitative segmentation results of the model trained on synthetic data only.

\textit{UPGen} sets a new state-of-the-art result for the CVPPP LSC as shown by the mean performance over the four test datasets (Table~\ref{tab_results_literature}). The two most recent and highest performing approaches, Ward~\etal~(2018) and Kuznichov~\etal~(2019), deploy the same instance segmentation architecture used in this paper and also employ synthetic data. The superior plant species generalisation of \textit{UPGen} is demonstrated here by the lower variance across datasets.

\input{tables/4_stateOfArt.tex}
\subsection{In the Wild}\label{sec_results_unseenData}

\input{tables/4_wild_contextBackground.tex}

In order to investigate the generalisation ability, we evaluate models trained on only synthetic data on two completely unseen datasets. Except where specified, no samples or parts (textures, leaf shapes) from these datasets were used for training these models. 
Table~\ref{tab_results_capsicum} compares using our domain randomised synthetic data to a common transfer learning approach, training on existing available datasets from a similar domain (all CVPPP datasets) and to \cite{ward2018deep}, the state-of-the-art method on the CVPPP-A1 dataset.
Further, we demonstrate a performance improvement when training on a new synthetic dataset; one with in context background textures as described in Section~\ref{sec_method_background_texture}.

\begin{figure}[b!]
	\centering
	\subfigure[Ground Truth]{
		\includegraphics[width=0.14\textwidth]{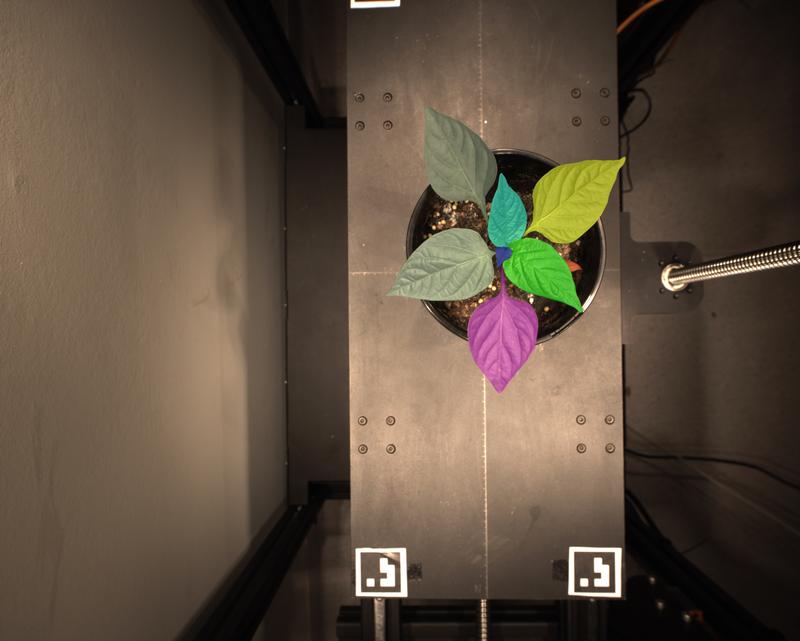}
	}
	\hspace{-2mm}
	\subfigure[Out of Context]{
		\includegraphics[width=0.14\textwidth]{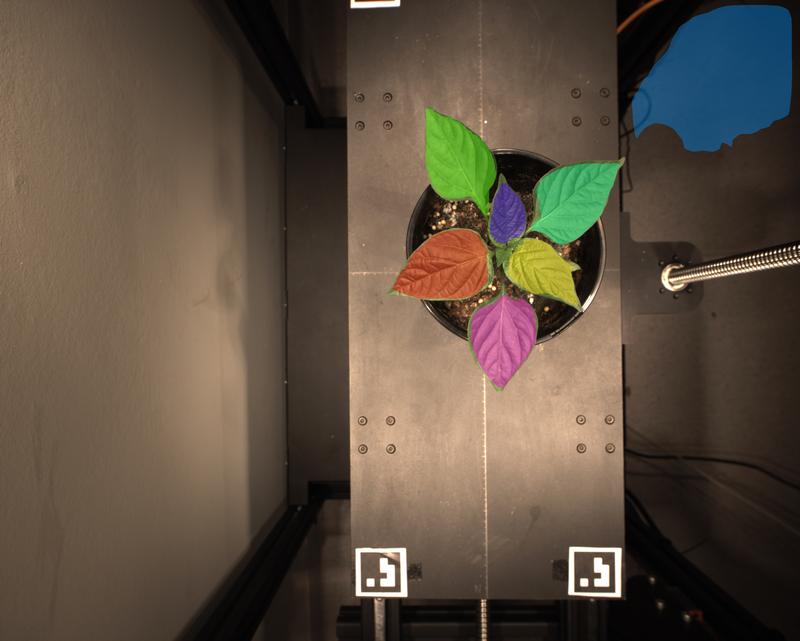}
	}
	\hspace{-2mm}
	\subfigure[In Context]{
		\includegraphics[width=0.14\textwidth]{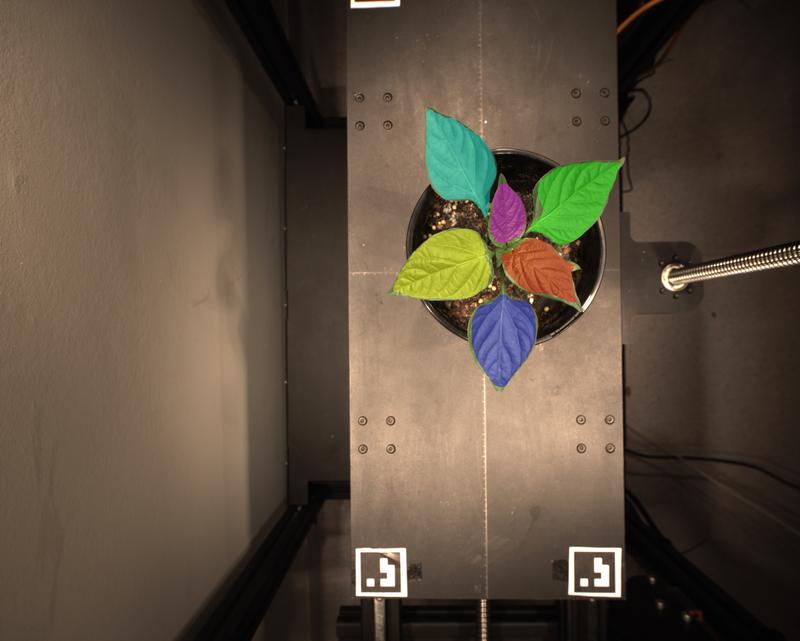}
	}
	\newline
	\subfigure[Ground Truth]{
		\includegraphics[width=0.14\textwidth]{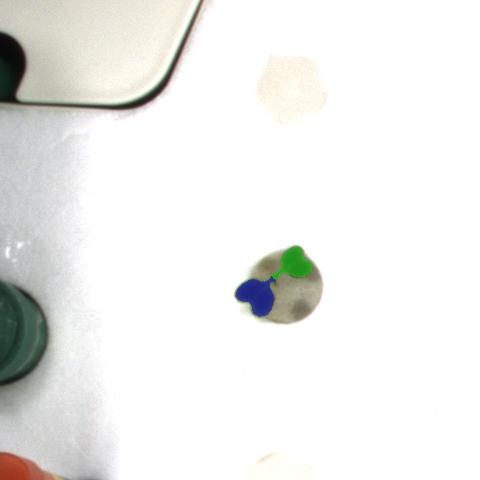}
	}
	\hspace{-2mm}
	\subfigure[Out of Context]{
		\includegraphics[width=0.14\textwidth]{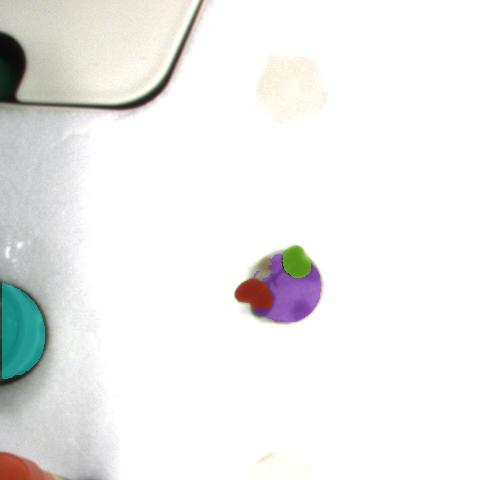}
	}
	\hspace{-2mm}
	\subfigure[In Context]{
		\includegraphics[width=0.14\textwidth]{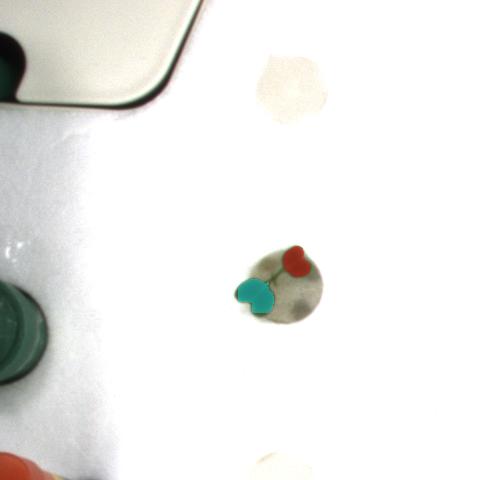}
	}
	\caption{Leaf segmentation results for models trained on synthetic data and synthetic data with in context background textures. A greater segmentation score is achieved when using synthetic data with in context background textures through reduced false positives which are seen here.}
	\label{fig_syntheticContextResults}
\end{figure}

A model trained on our synthetic data generalises to unseen plant species and imaging systems better than one trained on existing real data (CVPPP-All). In Table~\ref{tab_results_capsicum} our synthetic data (labeled \textit{out of context} outperformed transfer learning from the CVPPP data by 4.38\% and 39.63\% on the capsicum and Komatsuna test images respectively.

Our method also outperforms the previous state-of-the-art method, \cite{ward2018deep}. Using their pretrained model\footnote{\url{https://research.csiro.au/robotics/our-work/databases/synthetic-arabidopsis-dataset/}}, we benchmark their performance on the same Komatsuna and Capsicum datasets. The previous state-of-the-art, \cite{Kuznichov_2019_CVPR_Workshops}, were not available for comparison because they did not release their data or pretrained model. It is expected to perform similarly to \cite{ward2018deep} as their synthetic data is also constructed using the CVPPP datasets and, hence, species specific. Their method performs worse than transfer learning from real data on the capsicum data and 12.9\% worse than \textit{UPGen} on the Komatsuna dataset. Noting that \cite{ward2018deep} and \cite{Kuznichov_2019_CVPR_Workshops} employ the same instance segmentation architecture as used in this study, we attribute the performance improvements to the synthetic data used for training in each method. Specifically, the generalisation ability of \textit{UPGen} results from the ability to input many plant species' leaf geometries and textures and the modeling of plant trunks and leaf stems to produce a wider range of 3D plant models.

Further performance gains were achieved by training on the in context background texture synthetic data. In these experiments, new synthetic datasets were generated on background textures from the imaging systems in the test datasets. As described in section~\ref{sec_method_background_texture}, these images were obtained by inpainting. However, if this approach were used in a plant phenotyping facility, one could capture the background by photographing an empty plant pot. In Table~\ref{tab_results_capsicum} the in context synthetic data outperformed transfer learning from the CVPPP data by 26.06\% and 51.46\% on the capsicum and Komatsuna test images respectively.

These performance gains demonstrate the benefits of UPGen and simulated data methods over generative models. Such improvements were the result of swapping out the data textures to better match the desired application domain which, as discussed in the related work section, is not possible in GAN based synthetic data approaches~\citep{giuffrida2017arigan, Kuznichov_2019_CVPR_Workshops}.
The effect of using in context background for synthetic data is shown qualitatively in Figure~\ref{fig_syntheticContextResults} and quantitatively in Table~\ref{tab_results_capsicum}. Table~\ref{tab_results_capsicum_precision_recall} presents the precision and recall of the segmentation result in Table~\ref{tab_results_capsicum}. A greater increase in precision compared to the decrease in recall confirms that the segmentation improvement is a result of reduced false positive segmentations. To calculate this result; true positive segmentations were identified based on having an intersection over union (IoU) with the corresponding ground truth of at least $0.5$.
Further, the use of in context synthetic data resulted the greatest consistency across samples in the test datasets. This is shown by the lowest standard deviation in Table~\ref{tab_results_balanced_training} (bottom row).

\subsection{Dataset Size}\label{sec_results_datasize}
\begin{figure}[t!]
	\centering
	\includegraphics[width=0.45\textwidth]{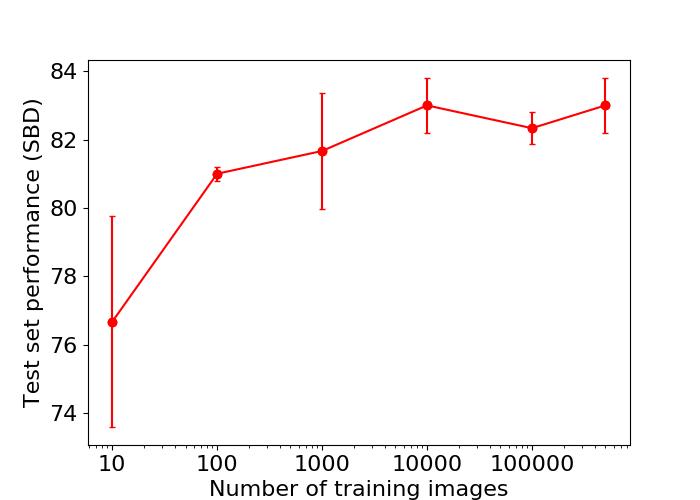}
    \caption{Model performance when training on different amounts of synthetic data. Comparable performance to the baseline is achieved at 1000 training images. For the baseline, a score of 86 SBD was achieved when training on 810 real data samples from the same distribution. Dataset size is shown on the $x$ axis using a logarithmic scale. Smaller datasets were sampled from the largest dataset. As performance is sensitive to the random choice of training samples, each data point represents the mean of three replicates. Performance is quoted on the CVPPP A5 test dataset, which is a combination of the A1-4 test datasets.}
    \label{fig_datasize}
\end{figure}

The influence of the number of synthetic training samples on the leaf instance segmentation performance was investigated by training models on datasets ranging from 10 to 500,000 samples. The size of the largest dataset was determined by the computational resources available. Figure~\ref{fig_datasize} presents the model performance on the CVPPP A5 dataset which contained multiple plant species. The standard deviation across the three replicates decreases for larger datasets as the performance is less dependent on specific data samples. Model performance starts to plateau when trained on datasets of at least 10,000 samples. This motivated the synthetic dataset size used for the other experiments in this paper and demonstrates the potential use of synthetic datasets in image-based plant phenotyping systems. The CVPPP dataset contains 810 training samples, specific to three plant species and four imaging scenarios. A model trained on this data (CVPPP-All in Table~\ref{tab_results_balanced_training}) achieves 86.0 SBD. It is expected that this is greater than the result shown here because the training and test data were drawn from the same distribution, the CVPPP dataset. The utility of \textit{UPGen} data is demonstrated by achieving 82.0 SBD when trained on a 1000 sythetic images, a similar sized dataset to the CVPPP dataset. Using \textit{UPGen}, one can produce significantly more training data for free and apply it to any plant species.

\subsection{Ablation Study}\label{sec_results_ablation}

\begin{figure}[b!]
	\centering
	\includegraphics[width=0.45\textwidth]{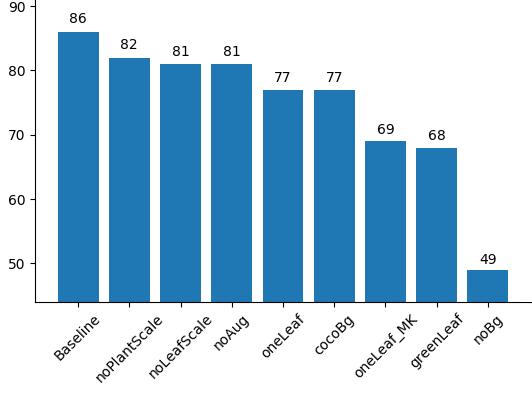}
    \caption{Impact on performance when ablating various aspects of the synthetic data or training procedure. Performance is quoted on the CVPPP A5 test dataset, which is a combination of the A1-4 test datasets.}
    \label{fig_ablation}
\end{figure}

An ablation study was conducted to investigate the contribution of each aspect/parameter of the \textit{UPGen} pipeline on the performance of a trained model. In each experiment a dataset of 10,000 images was used and the model was trained for 280,000 steps. Each model was evaluated on the CVPPP-A5 test dataset. This is a combination of all CVPPP test data, A1-4. It was selected as it is highly variable with images from four different plant species/mutations and imaging scenarios. Note that each plant species is not equally represented in this test dataset.
Figure~\ref{fig_ablation} presents the results of the ablation study, each omitted component is ordered by its effect on performance. The effect of these on segmentation performance is compared to the baseline. Each of the omitted or changed data components are described below.

\noindent \textbf{Baseline.} All ablations are compared to the baseline model which was trained on synthetic data only. It achieved 86 SBD on the test data.

\noindent \textbf{Out of context background (cocoBg).} Each data sample was synthesised using a randomly sampled image from the MS COCO dataset as the background. Here, the segmentation score is reduced to 77.0 SBD when training on data without in context backgrounds.

\noindent \textbf{No Background (noBg).} Each plant was rendered on a black background removing all background textures. This results in further performance reductions relating to changes in background textures compared to \textit{cocoBg}. This also had the greatest effect by a significant margin on the segmentation performance.

\noindent \textbf{No leaf texture (greenLeaf).} Each leaf is textured with a uniform shade of green. Rendering all leaves with the same texture reduced performance to 68.0 SBD.

\noindent \textbf{One context leaf (oneLeaf).} When limiting the randomisation in leaf geometry by using a single inspiration leaf (CVPPP dataset) from the same plant species as the test dataset, performance was reduced to 77.0 SBD.

\noindent \textbf{One out of context leaf (oneLeaf\_MK).} However, when using a single inspiration leaf from a plant species different to that in the test data (one from the MalayKew dataset), performance was further reduced to 69.0 SBD.

\noindent \textbf{No leaf scaling (noLeafScale) and No plant scaling (noPlantScale).} Disabling the random leaf sizing or plant sizing had a small effect. Omitting these reduced the segmentation performance to 81.0 and 82.0 SBD respectively.

\noindent \textbf{No augmentation (noAug).} Ablating the randomisation of each data sample by disabling augmentation during training also had a small effect on performance. This suggests that model robustness to \textit{domain gap} and \textit{species gap} is a result of domain randomisation more than data augmentation.

Conducting this ablation study also demonstrates the benefits of our proposed synthetic data solution over other generative approaches including GAN based methods. In algorithmically generated data like \textit{UPGen}, one can easily turn on or off, swap in or out and tweak dataset parameters. Benefits of such capabilities are clearly shown in the performance improvements achieved by changing the background textures in the data (Table~\ref{tab_results_capsicum}). Conversely, collecting specific data and retraining would be required to achieve similar capabilities and performance improvements using a GAN based synthetic data approach. The state-of-the-art GAN architectures have developed since the proposal of ARIGAN, a synthetic plant image GAN~\citep{giuffrida2017arigan}. Recent work by \cite{bau2019gandissect} is a step towards parameterised synthetic data using GANs. The developed a method to interactively manipulate objects in a scene outputted by a GAN. Applying such a method to plant data would still face challenges producing a corresponding image segmentation label. Further, through the application of domain randomisation, \textit{UPGen} can produce a more widely distributed dataset for training than a GAN method. A comparison between \textit{UPGen} and ARIGAN~\citep{giuffrida2017arigan} data could not be shown empirically because no image segmentation labels were produced by ARIGAN. In our previous work, we have demonstrated a wider than real world data distribution achieved using algorithmically generated data~\citep{ward2018deep, ward2018synthetic}. \cite{bau2019seeing} explored what GANs can and cannot generate. The results noted that their model did not generate enough pixels of complex objects including people, trees, or signboards compared to the training distribution.

%% file: tables/4_batchBalance.tex
\begin{table}[t!]
\caption{Instance segmentation results on the CVPPP test datasets. Synthetic data (\textit{UPGen}) is a valid alternative to real training data (CVPPP-All). Further, combining real and synthetic data results in further improvements. Batch balancing performs marginally better on A1 while fine tuning achieved comparable or better performance on A2, A3 and A4.}
\label{tab_results_balanced_training}
\begin{center}
\begin{tabular}{|l|p{0.6cm}|p{0.6cm}|p{0.6cm}|p{0.6cm}||p{1.4cm}|}%
\hline
Training & \multicolumn{5}{|c|}{Segmentation Score (SBD)} \\
\hline
 & A1 & A2 & A3 & A4 & Mean\\
 \hline
CVPPP-All & 81.0 & 81.0 & 84.0 & 86.0 & 83.0 (2.5)\\
UPGen & 86.0 & 81.0 & 83.0 & 86.0 & 84.0 (2.5)\\ %
\hline
\hline
Batch Balanced & \textbf{90.0} & 85.0 & 81.0 & 87.0 & 85.8 (3.8)\\
Fine Tuning & 89.0 & \textbf{88.0} & \textbf{86.0} & \textbf{88.0} & \textbf{87.8 (1.3)}\\
\hline
\end{tabular}
\end{center}
\end{table}

%% file: tables/4_stateOfArt.tex
\begin{table*}[!htpb]
\caption{Our best performing models trained compared to the state of the art. Our best performance was achieved when combining real and synthetic data during training. Results are presented column-wise for each data subset in the CVPPP test data. The mean column presents the mean and standard deviation across the datasets, A1-4. Note that~\cite{ward2018deep},~\cite{Kuznichov_2019_CVPR_Workshops} and UPGen all used the same implementation of the Mask-RCNN architecture.}
\label{tab_results_literature}
\centering
\begin{tabular}{|l|c|c|c|c|c|}%
\hline
\multirow{2}{*}{Method} & \multicolumn{4}{|c|}{CVPPP test set (SBD)} & \multirow{2}{*}{Mean (SD)} \\
\cline{2-5}
 & A1 & A2 & A3 & A4 & \\ 
\hline
RIS + CRF (\cite{romera2016recurrent}) & 66.6 & - & - & - & - \\ 
MSU (\cite{scharr2016leaf}) & 66.7 & 66.6 & 59.2 & -& - \\ 
Nottingham (\cite{scharr2016leaf}) & 68.3 & 71.3 & 51.6 & -& - \\
Wageningen (\cite{yin2014multi}) & 71.1 & 75.7 & 57.6 & -& - \\
IPK (\cite{pape20143}) & 74.4 & 76.9 & 53.3 & -& - \\
\cite{cSalvadore} & 74.7 & - & - & -& - \\
\cite{kulikov2018instance} & 80.4 & - & - & -& - \\
\cite{de2017semantic} & 84.2 & - & - & -& - \\
\cite{ren2017end} & 84.9 & - & - & - & - \\
\cite{kulikov2019instance} & 89.9 & - & - & -& - \\
\cite{ward2018deep} & \textbf{90.0} & 81.0 & 51.0 & 88.0 & 77.5 (15.7)\\
\cite{Kuznichov_2019_CVPR_Workshops} & 88.7 & 84.8 & 83.3 & \textbf{88.6} & 86.4 (2.73) \\
UPGen (Ours) & \textbf{90.0} & \textbf{88.0} & \textbf{86.0} & 88.0 & \textbf{88.0 (1.63)}\\ %
\hline
\end{tabular}

\end{table*}

%% file: tables/4_wild_contextBackground.tex
\begin{table}[b!]
\caption{Leaf instance segmentation results on test datasets of completely unseen plant species and imaging scenarios. A model trained on our synthetic data is shown to better transfer to unseen data than one trained on existing publicly available real data (CVPPP-All). A further generalisation boost is seen when training on a custom synthetic dataset containing \textit{in context} background textures.}
\label{tab_results_capsicum}
\begin{center}
\begin{tabular}{|l|c|c|}
\hline
Training Data & \multicolumn{2}{|c|}{Segmentation Performance (SBD)} \\
\hline
 & Capsicum & Komatsuna \\
\hline
CVPPP-All & 72.49 (7.15) & 51.34 (16.15) \\
\cite{ward2018deep} & 64.57 (10.83) & 62.43 (17.04) \\
\textit{UPGen} & 75.67 (9.52) & 71.69 (16.76) \\
\textit{UPGen} - In context & \textbf{91.38 (2.27)} & \textbf{77.76 (15.26)} \\ 
\hline
\end{tabular}
\end{center}
\end{table}

\begin{table*}[!htpb]
\caption{Precision and recall metrics for the leaf instance segmentation results on the unseen test sets. Greater changes in precision than recall confirm that the performance improvements achieved by using in context synthetic data (Table~\ref{tab_results_capsicum}) are a result of reduced false positives. True positive segmentations were calculated at an IOU of 0.5.}
\label{tab_results_capsicum_precision_recall}
\begin{center}
\begin{tabular}{|l|c|c|c|c|}
\hline
Training Data & \multicolumn{2}{|c|}{Precision (IOU=0.5)} & \multicolumn{2}{|c|}{Recall (IOU=0.5)}\\
\hline
 & Capsicum & Komatsuna & Capsicum & Komatsuna\\
\hline
Out of context & 81.76 & 71.24 & 74.71 & 86.63\\ 
In context & \textbf{99.19} & \textbf{78.63} & \textbf{69.89} & \textbf{85.98}\\
\hline
Percent Change & +21.32 & +10.37 & -6.45 & -0.75\\
\hline
\end{tabular}
\end{center}
\end{table*}

%% file: chapters/5_discussion.tex
\section{Conclusion}\label{sec_discussion}
In this paper, we proposed the problem of \textit{species gap} and the implications it has on the use of deep learning algorithms for image-based plant phenotyping. With many different plant species in existence, growing plants, imaging them and then labeling the amounts of data required for modern deep learning algorithms is not tractable. We proposed \textit{UPGen} as a synthetic data solution to the \textit{species gap} challenges in precision agriculture. In the design and validation of our method, we explored the use of synthetic data for plant imaging. We focused on informing the process of using deep learning image analysis to automate plant phenotyping measurements.
In \textit{UPGen} synthetic 3D plant models are algorithmically assembled and then synthetic data samples are then rendered. This approach avoids the time and cost overhead of data collection and annotation. It is also directly applicable to high throughput imaging and plant phenotyping setups where the image background does not change. The proposed method leverages quantity over quality when it comes to data realism. Our methods incorporate biological mutations and stochasticity through domain randomisation. In our experiments and discussion we demonstrate the benefits of our proposed method over common synthetic data approaches such as GANs or cut and paste imaging.
Notably, we presented segmentation results on two unseen datasets of different plant species, capsicum and komatsuna. Our approach outperformed the baseline (transfer learning from the CVPPP datasets) by 26.06\% and 51.46\% on the capsicum and komatsuna test datasets respectively. We achieved significant performance improvements by having fine control over different aspects of the generated synthetic data. For example, that experiment and the ablation study of the different parameters of our synthetic data revealed the background texture to have the most significant effects on leaf instance segmentation performance. The background texture is a data attribute which can change between green houses, growth mediums for different species of plants or across different plant imaging facilities. Where our approach can simply tweak this parameter, a GAN based one would required retraining.
We also validate our method by by competing in the CVPPP leaf segmentation challenge (LSC). State-of-the-art performance was achieved when training on both real and synthetic data. The same model was used for each test dataset, containing different plant species, showing that it addressed the \textit{species gap} and \textit{domain gap} well.
Training on synthetic data alone, we achieved comparable performance to the baseline of training on real data from the same distribution as the test images. This demonstrated that our data was realistic and that the \textit{domain gap} was small. Despite being able to generate a training dataset of infinite samples, we achieve comparable results to training on real data with as few as 1000 synthetic images.

This paper focused on the task of leaf instance segmentation to demonstrate the effectiveness of synthetic data in automating phenotypic measurements such as leaf area. We hope this work encourages new lines of research in the use of deep learning and computer vision for image-based plant phenotyping without the costs of physical experimentation, data collection and annotation. The \textit{UPGen} pipeline outputs several modalities in addition to the RGB and instance segmentation labels used in this study. These include the 3D plant model, a depth map, plant segmenation label, leaf center location, leaf count and the skeleton graph describing the plant structure. \cite{zhou2017leaf} and \cite{sa2017peduncle} have investigated applications of plant skeletal structure and the use of depth information for plant growth measurement and fruit picking on real plant images respectively. \cite{sa2017weednet} improved weed classification performance by making use of multi-spectral imaging. Future work will investigate simulating such modalities for synthetic data.

\section*{Acknowledgements}
The authors would like to thank the members of the Computer Vision Problems in Plant Phenotyping (CVPPP) group for forming this research community and making their data available. We also would like to thank Eranda Tennakoon and Josh Knights for their feedback. The work was supported by the AgTech Cluster for Robotics and Autonomous Systems, CSIRO.

%% file: chapters/8_appendix.tex
\section{UPGen: The Universal Plant Generator}\label{appendix_upgen}
\subsection{Leaf Geometry Processing}\label{appendix_leaf_geometry}
\begin{figure}[b!]
	\centering
	\includegraphics[width=0.45\textwidth]{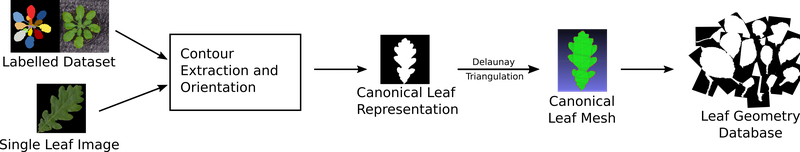}
    \caption{The process of extracting inspiration leaf geometry from existing real datasets.}
    \label{fig_leaf_extraction_pipeline}
\end{figure}

Inspiration leaf geometries were obtained from existing data. Each leaf was converted to a canonical representation to ensure consistent inputs to the leaf insertion pipeline. Figure~\ref{fig_leaf_extraction_pipeline} illustrates the extraction and conversion to canonical represenation procedure. Leaves were extracted from the CVPPP A1-4~\citep{scharr2016leaf} datasets; MalayaKew~\citep{lee2015deep} dataset and Migicovsky's~\etal~\citep{migicovsky2018morphometrics} apple leaf dataset (refered to as appleLeaf).
The canonical leaf representation consisted of the leaf being masked, arranged vertically (stem down) and scaled to a consistent height. A planar mesh defining the leaf geometry was then obtained by employing the Delaunay triangulation algorithm.

Images from the MalayaKew and appleLeaf datasets contained a single leaf. To obtain canonical leaf masks, a threshold and morphological operations were applied to each image in order to isolate the leaf. To obtain the canonical orientation, a minimum area rectangle was fitted to each leaf. The centre of the rectangle was compared to the leaf centre of mass (CoM). Then, the leaf was rotated such that the two points were vertically aligned and the CoM was above the rectangle centre. This was based on the assumption that the majority of leaf pixels contributing to the leaf CoM were in the blade and not the stem while the rectangle described the absolute centre. The same method was applied to the two datasets however an initial orientation angle guess provided with each MalayaKew sample was also used.

Each image in the CVPPP datasets contained a single plant and, hence, multiple leaves. Each leaf was identified using the leaf instance segmentation masks provided with the dataset. The plant centre was computed to be the CoM of the plant mask. A contour was fitted to each leaf and the tip and stem were then computed to be the vertexes of greatest and smallest euclidean distance from the plant centre respectively. 
By vertically aligning the leaf stem and tip, canonical orientation was achieved.

As previously mentioned, a planar mesh was then computed for each inspiration leaf in canonical form. 
In our experiments we used the described methods to assemble 17,957 leaves to produce training images, 186 were kept out for test images. These were compiled from 46 different plant species.

\subsection{Leaf Texture Processing}\label{appendix_leaf_texture}
Multiple sources were utilised to assemble a set of leaf textures. Namely the low frequency content textures from~\citep{ward2018deep}, high frequency content textures (containing strong edges such as leaf veins) from open license texture databases and texture patches from the leaves of the MalayaKew dataset. Texture patches were rectangular image patches extracted from within the border of a leaf and, hence, containing only leaf texture.
Following this, we used 40,552 textures to produce synthetic training images and held out 2,964 for test images.\\

To add further variance in the set of leaf textures, augmentations were applied to the images. The leaf texture augmentation sequence consisted of the following operations, the probability of each augmentation being applied is presented in the parentheses: flip left-right (0.5); flip up-down (0.5); randomly crop or pad each side of the image up to 25\% of the dimension (0.5); apply an affine transform (0.5); replace the texture with it's superpixels (0.1); apply a Gaussian blur with sigma between 0 and 3 (1.0); randomly add between -10 and 10 to the image brightness, hue and saturation independently (1.0) and apply a perspective transform of scale between 0.01 and 0.1 (0.5). The random affine transform consisted of: scaling the image by a constant between 0.8 and 1.2; translating the image in both directions independently between 0 and 20\% of the axis dimension; rotating the image by an angle between -45 and 45 degrees and shearing the image by an angle between -15 and 15 degrees. The order in which augmentations were applied to each image was random.

\subsection{Background Texture Processing}\label{appendix_background_texture}
\begin{figure}[b!]
	\centering
	\includegraphics[width=0.45\textwidth]{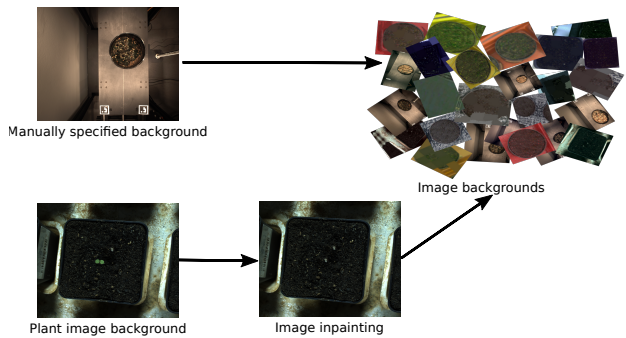}
    \caption{The process of obtaining a set of background textures.}
    \label{fig_background_texture}
\end{figure}

In controlled imaging scenarios, such as image-based plant phenotyping platforms~\citep{scharr2016leaf, moghadam2017plant}, the background remains fairly consistent. Following~\citep{ward2018deep} and~\citep{Kuznichov_2019_CVPR_Workshops}, in context backgrounds were obtained from existing images by inpainting the visible plant. Figure~\ref{fig_background_texture} visualises the steps of inpainting to obtain a background image. Note that an alternative to inpainting, when one has access to the relevant plant imaging platform, is to photograph the background before placing a plant pot there.
In our experiments, we generated synthetic data using 18 background textures consisting of imaged soil and inpainted images from the CVPPP datasets.

Similar to the leaf textures, image augmentations were applied to the background textures to add further variance. As previously mentioned, background textures were observed to remain fairly consistent in imaging scenarios. To model this, the augmentations applied to background textures were tame compared to those described above. The augmentations differing from the leaf texture process were, the chance of each augmentation being applied is presented in the parentheses: randomly crop or pad each side of the image up to 5\% of the dimension (0.5) and to random additions were made to the image hue or saturation. Further the random affine transform parameters were: scaling the image by a constant between 0.8 and 1.2; translating the image in both directions independently between 0 and 10\% of the axis dimension; rotating the image by an angle between -5 and 5 degrees and shearing the image by an angle between -3 and 3 degrees. As in the leaf texture pipeline, augmentations were applied to each image in a random order.

\subsection{Implementation Details}\label{appendix_implementation_details}
\textit{UPGen} was designed and implemented using the Python scripting interface in Blender (v2.79b). Here, processing and permuting the different inputs and parameters of \textit{UPGen} is performed. The 3D plant model is assembled by algorithmically sampling a leaf, positioning it in free space and connecting it to the plant by computing the stem location and dimensions.
We render synthetic images from the 3D plant model placed on a randomly sampled background texture. The Cycles renderer with Lambertian and Oren-Nayar diffuse reflection shading is used. Esing the native Blender renderer is used to produce the segmentation labels. To ensure the integrity of the segmentation labels (\textit{i.e.} one RGB value per leaf) all shading and anti-aliasing is disabled.

\section{Leaf Instance Segmentation Experiments}
\subsection{Performance Metric}\label{appendix_dice_definition}
Comparison and measurement of the leaf instance segmentation is reported using the \emph{symmetric best dice} ($SBD$) metric, it's formulation is described here. It is the metric used for the CVPPP leaf segmentation competition~\citep{scharr2016leaf} and, hence, allows direct comparison of our results to existing literature. The \emph{S{\o}rensen-Dice similarity coefficient} ($DSC$) is a set overlap statistic (Equation~\ref{eqn_dice}) which can be applied to binary segmentation masks.

\begin{equation}
    DSC (\%) = \frac{2|P^{gt} \cap  P^{pred}|}{|P^{gt} | +  |P^{pred}|}
    \label{eqn_dice}
\end{equation}

\emph{Best dice} (BD) extends this to instance segmentation masks. Here the instance segmentation mask from ground truth set B which yields the largest $DSC$ is the one compared to each segmentation from set A. Note that this is computed independently for each predicted instance segmentation mask so it is possible that multiple predicted masks are compared to the same ground truth instance mask. Best dice is presented in Equation~\ref{eqn_bd}:

\begin{equation}
	BD(A, B) = \frac{1}{M} \sum_{i=1}^{M} \max\limits_{1 \leq j \leq N} \frac{2|A_i \cap B_j|}{|A_i| + |B_j|}
    \label{eqn_bd}
\end{equation}

where $A$ and $B$ correspond to the set of $M$ and $N$ segmentations respectively. A further extension to this is the \emph{symmetric best dice} ($SBD$). Which computes the minimum score between ground truth and prediction permutations to account for the aforementioned limitations of $BD$. It is presented in Equation~\ref{eqn_sbd} and used in all experiments in this paper: 

\begin{equation}
	SBD(S^{pred}, S^{gt}) = min(BD(S^{pred}, S^{gt}), \hspace{1mm} BD(S^{gt}, S^{pred}))
    \label{eqn_sbd}
\end{equation}

where $S^{x}$ represents the set of instance segmentation masks, ground truth or predicted. Results are presented as the mean $SBD$ across all samples in the test dataset.

%% file: chapters/7_visuals.tex
\begin{figure*}[!htpb]
	\centering
	\includegraphics[height=0.9\textheight]{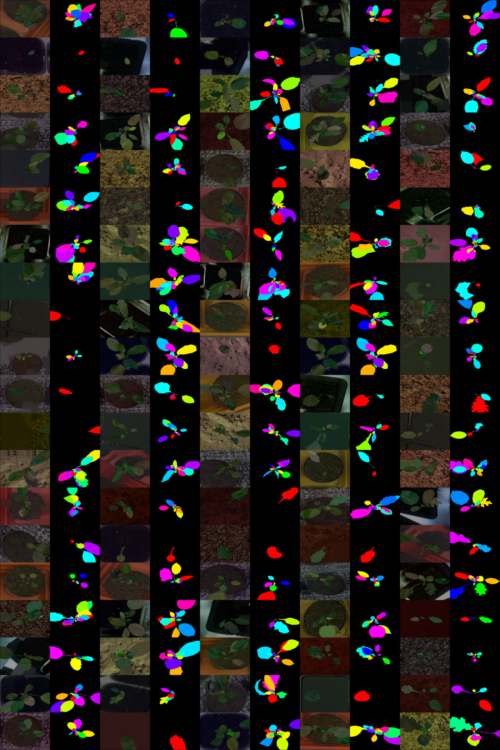}
    \caption{Example synthetic plant images generated by the \textit{UPGen} pipeline. Five column pairs present example RGB images (left column) and the corresponding ground truth per-pixel leaf instance annotations (right column).}
    \label{fig_dataVisual}
\end{figure*}